%% file: main.tex
\documentclass[letterpaper, preprint, paper,11pt]{AAS}	

\usepackage{bm}
\usepackage{amsmath}
\usepackage{subfigure}
\usepackage[colorlinks=true, pdfstartview=FitV, linkcolor=black, citecolor= black, urlcolor= black]{hyperref}
\usepackage{overcite}
\usepackage{footnpag}			      	

\usepackage{xfrac}
\usepackage{mathdots}
\usepackage{relsize}
\usepackage{subcaption}
\usepackage{mathtools, nccmath}
\usepackage{amsfonts}
\usepackage{acronym}

\usepackage{array}

\newacro{SCvx}{Successive Convexification}
\newacro{SCP}{Sequential Convex Programming}
\newacro{DRO}{Distant Retrograde Orbit}
\newacro{NRHO}{Near-Rectilinear Halo Orbit}
\newacro{CRTBP}{Circular Restricted Three Body Problem}
\newacro{CRLB}{Cramer-Rao Lower Bound}
\newacro{RMS}{Root Mean Square}
\newacro{SSA}{Space Situational Awareness}
\newacro{SDA}{Space Domain Awareness}
\newacro{DSN}{Deep Space Network}

\PaperNumber{XX-XXX}

\begin{document}

\title{Information-Based Trajectory Planning for Autonomous Absolute Tracking in Cislunar Space}

\author{Trevor N. Wolf\thanks{Ph.D. Candidate, Department of Aerospace Engineering and Engineering Mechanics, and AIAA Student Member.},  
Brandon A. Jones\thanks{Associate Professor, Department of Aerospace Engineering and Engineering Mechanics, and AIAA Associate Fellow.}
}

\maketitle{}

\begin{abstract}

The resurgence of lunar operations requires advancements in cislunar navigation and Space Situational Awareness (SSA). Challenges associated to these tasks have created an interest in autonomous planning, navigation, and tracking technologies that operate with little ground-based intervention. This research introduces a trajectory planning tool for a low-thrust mobile observer, aimed at maximizing navigation and tracking performance with satellite-to-satellite relative measurements. We formulate an expression for the information gathered over an observation period based on the mutual information between augmented observer/target states and the associated measurement set collected. We then develop an optimal trajectory design problem for a mobile observer, balancing information gain and control effort, and solve this problem with a Sequential Convex Programming (SCP) approach. The developed methods are demonstrated in scenarios involving spacecraft in the cislunar regime, demonstrating the potential for improved autonomous navigation and tracking.
\end{abstract}

\section{Introduction}
\input{introduction}


\section{Problem Formulation}
\input{problem_statement}

\section{Sequential Convex Programming for Trajectory Generation}
\input{convex_programming}


\section{Convex Objective Approximation}
\input{convex_objective}


\section{Numerical Case Studies}
\input{numerical_cases}


\section{Conclusion and Future Work}
\input{conclusion}

\bibliographystyle{AAS_publication}   
\bibliography{references}   

\end{document}

%% file: introduction.tex
The resurging interest in lunar operations creates new requirements for cislunar navigation and \ac{SSA}. As multiple actors eagerly prepare to claim a stake in this domain, like Earth orbit, cislunar space will undoubtedly become a congested and contentious domain. The region between the Earth and the Moon occupies a volume orders of magnitude larger than that Earth orbiting space objects populate. Moreover, the dynamical behavior of space objects in this environment exhibits strong nonlinearities, and small perturbations can easily depart a trajectory from its expected path. While the dynamical instability can be leveraged to efficiently transfer between cislunar regions, say with low-thrust electric propulsion systems, it exacerbates difficulties in uncertainty quantification over extended measurement gaps \cite{Wolf_2021, Jones_2024}. Adding to this, accurate ground-based tracking is only tangible with exquisite facilities, such as the \ac{DSN}. The anticipated operational growth in the cislunar environment will certainly outpace the capacity of these already over-taxed legacy systems. To address these challenges, the spaceflight community has an expressed interest in developing autonomous cislunar tracking and navigation technologies that can operate with minimal ground-based intervention. Motivated by this objective, our work introduces a trajectory planning tool, compatible with a low-thrust mobile observer. The tool is designed to maximize an observer's absolute navigation and tracking performance at a fixed control effort, based on satellite-to-satellite relative measurements.

The work of Markley\cite{Markley_1984} is the first to propose the concept of inter-satellite measurements for absolute navigation with two spacecraft in Earth orbit. Such a navigation system is analogous to an experiment that measures the difference of two (or more) accelerometers at a very long baseline \cite{Psiaki_1999}. In this analogy, the measurement is depended on a $1/r^2$ force law attributed to Earth, and thus, the navigation state of the two objects. Twice-differencing a sequence of inter-satellite relative positions provides an equivalent measurement to this analogy. Generally, increasing the baseline of the test masses provides a higher degree of differentiation, and therefore, increases the navigation accuracy. Follow-on work to Markley's extended this concept to absolute navigation in cislunar space using crosslink range measurements \cite{Hill_2008}, and more recently, optical measurements \cite{Greaves_2021}. 

Reconfiguring a mobile sensor platform can significantly increase its navigation and tracking accuracy. The predominant mode in performance gain is generated by increasing the baseline between the observer and the test masses, e.g., the targets. However, an optimal reconfigured trajectory course is determined by both the measurement type, and the expected dynamics of the observer and the targets. In the context of similar problems, works, such as that by Fawcet\cite{Fawcet_1988}, evaluates the effect of course maneuvers on state estimation performance using a \ac{CRLB} analysis. Within the spaceflight community, other studies have developed analytic formulations for optimal impulsive maneuvers based on geometric definitions of observability in Earth orbit\cite{Woffinden_2009, Grzymisch_2014}. Alternatively, information-based measures that directly relate to the performance of a sequential estimator have been applied in the information fusion literature. Among these, the Fisher information is most notable \cite{Hammel_1989, Oshman_1999, Hou_2021}. Similar to our problem at hand, recent work has studied the information collection problem for range determination based on optical measurement in the context of station keeping in the cislunar regime \cite{Greaves_2023}.  

This study extends our previous work \cite{Wolf_2023} and makes the following primary contributions. First, we develop an expression for the information gain that is consistent with the problem of simultaneously estimating the observer state, and an arbitrary number of catalog targets. Our expression is based on the \textit{mutual information} between a sequence of augmented states and the associated measurement set collected over the course of an observation interval. Secondly, we formulate an optimal trajectory design problem, compatible with a low-thrust mobile observer, that optimizes an objective that is a convex combination of the information gain, and the control effort exercised. The subsequent problem is solved with a \ac{SCP} approach. As is discussed in a later section, for compatibility in a convex program solver, the nonlinear optimization objective must be linearized at each solver iteration. We derive a first order approximation of our mutual information objective for this task. The methods we develop are applied to two different scenarios that consider spacecraft operating in cislunar space.

%% file: problem_statement.tex
\subsection{Optimal Control Problem}

This section describes the mathematical framework used in this study. We are interested in formulating an approach for trajectory generation that maximizes the information gain of a mobile observer, while limiting the control effort applied. In many cases, such as ours, these two aims are in conflict, so this task can be viewed as a multi-objective optimization problem. We choose to design a solution using as a fixed time, two-point boundary value optimal control problem of the form 
\begin{subequations}
    \begin{flalign}
        &\hspace{35ex} \min_{\boldsymbol{u}_{\mathcal{S}}(t)} J (\boldsymbol{x}_{\mathcal{S}}(t), \boldsymbol{u}_{\mathcal{S}}(t))&\label{eqn:original_cost}\\
        &\hspace{36ex}\mbox{s.t.} \hspace{1ex} \dot{\boldsymbol{x}}_{\mathcal{S}}(t) = \boldsymbol{f}(\boldsymbol{x}_{\mathcal{S}}(t), \boldsymbol{u}_{\mathcal{S}}(t), t)\mbox{,}&\\
        &\hspace{40ex} \boldsymbol{g}_{\mathrm{ic}}(\boldsymbol{x}_{\mathcal{S}}(t_0)) = \boldsymbol{0}\mbox{,}&\\
        &\hspace{40ex} \boldsymbol{g}_{\mathrm{tc}}(\boldsymbol{x}_{\mathcal{S}}(t_f)) = \boldsymbol{0}\mbox{,}&\\
        &\hspace{40ex} \|\boldsymbol{u}_{\mathcal{S}}(t)\|_2 \leq a_{\text{max}}\mbox{.}\label{eqn:nonlinear_thrust_constraint}&
    \end{flalign}
    \label{eqn:general_noncovex_continuous_problem}%
\end{subequations}
In the above, $\boldsymbol{x}_{\mathcal{S}}(t) \in \mathbb{R}^{n_x}$ and $\boldsymbol{u}_{\mathcal{S}}(t) \in \mathbb{R}^{n_u}$ are continuous time representations of the observer's state and control input, respectively. The boundary constraints are defined such that 
\begin{subequations}
    \begin{flalign}
        \boldsymbol{g}_{\mathrm{ic}}(\boldsymbol{x}_{\mathcal{S}}(t_0)) &= \boldsymbol{x}_{\mathcal{S}}(t_0) - \boldsymbol{x}_{\mathcal{S}, \text{ref}}(t_0)\mbox{,}\\
        \boldsymbol{g}_{\mathrm{tc}}(\boldsymbol{x}_\mathcal{S}(t_f)) &= \boldsymbol{x}_\mathcal{S}(t_f) - \boldsymbol{x}_{\mathcal{S}, \text{ref}}(t_f)\mbox{,}
    \end{flalign}
    \label{eqn:boundary_conditions}%
\end{subequations}
where $t_0$ and $t_f$ are the initial and terminal times, and $\boldsymbol{x}_{\mathcal{S}, \text{ref}}(\cdot)$ is a reference trajectory of interest. Thrust acceleration is bounded through Eq.~\ref{eqn:nonlinear_thrust_constraint}. We assume that the vehicle mass is constant through the entire trajectory arc, so $a_{\text{max}}$ is constant.

The problem objective is chosen as a convex combination of the control effort exercised and an information gain functional:  
\begin{equation}
    J(\boldsymbol{x}_{\mathcal{S}}(t), \boldsymbol{u}_{\mathcal{S}}(t)) =   \int_{0}^{t_f} \left[(1 - \alpha_h)\|\boldsymbol{u}_\mathcal{S}(t)\|_2 + \alpha_h \mathcal{R}\left(\boldsymbol{x}_{\mathcal{S}}(t), \boldsymbol{x}^{(1)}_\mathcal{T}(t), \cdots, \boldsymbol{x}^{(N_{\mathcal{T}})}_\mathcal{T}(t)\right)\right] dt.
    \label{eqn:objective_general form}
\end{equation}
Here, the information gain functional $\mathcal{R}(\cdot)$ is influenced by both the observer state, and a catalog set of target states, $\{\boldsymbol{x}_{\mathcal{T}}^{(i)}(t)\}_{i = 1}^{N_\mathcal{T}}$, where $N_{\mathcal{T}}$ is the number of targets. The parameter $\alpha_h \in [0, 1]$ is a homotopy that weighs the relative importance of the two competing objectives; the control input and the information gain. Later on, we show that solving the optimal control problem across a grid of $\alpha_h$ allows us to create a Pareto curve of optimal solutions that balance the two objectives.  

\subsection{Information-Driven Objective}

The aforementioned continuous description of our information-driven reward is convenient in formulation. In actuality, however, sensor measurements are collected recursively at discrete times. Moreover, for simplicity, we assume that the observer spacecraft does not maneuver between sequential measurements, collected over the course of an observation period. With these assumptions, the translational dynamics of the targets and observer between discrete time measurements are  
\begin{equation}
    \boldsymbol{x}_{\mathcal{T}, k + 1}^{(i)} = \boldsymbol{f}_k^{(i)}(\boldsymbol{x}_{\mathcal{T}, k}^{(i)}) + \boldsymbol{v}_{\mathcal{T}, k}^{(i)}\mbox{,}\hspace{0.25cm} \text{for } i = {1, 2, \cdots, N_\mathcal{T}},\label{eqn:target_dynamics_evolution}
\end{equation}
and,
\begin{equation}
    \boldsymbol{x}_{\mathcal{S}, k + 1} = \boldsymbol{f}_k(\boldsymbol{x}_{\mathcal{S}, k}) + \boldsymbol{v}_{\mathcal{S}, k}^{(i)},
\end{equation}
respectively. The discrete time translational dynamics are determined by integrating the continuous time equations of motion. The process noise, $\boldsymbol{v}_k \sim \mathcal{N}(\cdot; \boldsymbol{0}, Q_k)$, where $Q_k$ is the process noise covariance matrix integrated between time steps $k$ and $k + 1$. We assume that measurements of all targets are obtained synchronously, and each is provided by 
\begin{equation}
    \boldsymbol{y}_{k}^{(i)} = \boldsymbol{h}(\boldsymbol{x}_{\mathcal{T}, k}^{(i)}, \boldsymbol{x}_{\mathcal{S}, k}) + \boldsymbol{w}_k^{(i)}\mbox{,}\hspace{0.25cm} \text{for } i = {1, 2, \cdots, N_\mathcal{T}}.
\end{equation}
The measurement noise $\boldsymbol{w}_k \sim \mathcal{N}(\cdot; \boldsymbol{0}, R)$, where $R$ is the measurement noise covariance matrix. Our parameter of interest is the augmented state vector, 
\begin{equation}
    \boldsymbol{x}_{\mathcal{A}, k} = \left[ \boldsymbol{x}_{\mathcal{S}, k}^\top, \boldsymbol{x}_{\mathcal{T}, k}^{(1) \top}, \cdots,  \boldsymbol{x}_{\mathcal{T}, k}^{(N_{\mathcal{T}}) \top} \right]^\top,
\end{equation}
which concatenates the observer's state and those of the catalog targets. Similarly, we can define an associated augmented measurement vector as 
\begin{equation}
    \boldsymbol{y}_{\mathcal{A}, k} = \left[\boldsymbol{y}_{\mathcal{T}, k}^{(1) \top}, \cdots,  \boldsymbol{y}_{\mathcal{T}, k}^{(N_{\mathcal{T}}) \top} \right]^\top.
\end{equation}

The reward function is defined by the \textit{mutual information} between the augmented state and augmented measurement vectors. For now, consider one measurement collected at time step $k$. The mutual information for this single trial is
\begin{equation}
    \mathcal{I}_k(\boldsymbol{x}_{\mathcal{A}, k}; \boldsymbol{y}_{\mathcal{A}, k}) = \mathcal{D}_{\text{KL}}\left[p(\boldsymbol{x}_{\mathcal{A}, k}, \boldsymbol{y}_{\mathcal{A}, k})\|p(\boldsymbol{x}_{\mathcal{A}, k})p(\boldsymbol{y}_{\mathcal{A}, k})\right]\mbox{,}
        \label{eqn:MI_joint_KLD}
\end{equation}
where $\mathcal{D}[\cdot]$ is the Kullback-Leibler (KL) divergence, defined as 
\begin{equation}
    \mathcal{D}_{\text{KL}}[p(\xi)||q(\xi)] =  \int_{-\infty}^\infty p(\xi) \ln\left(\frac{p(\xi)}{q(\xi)}\right) dx\mbox{,}
\end{equation}
for arbitrary continuous distributions $p(x)$ and $q(x)$. The distribution $q(x)$ is assumed to have full support. The KL divergence is commonly interpreted as a measure of dissimilarity between probability distributions, and by marginalizing the joint distribution in Eq.~\ref{eqn:MI_joint_KLD}, we can re-write this quantity as
\begin{equation}
   \begin{multlined}[b]
    \mathcal{I}_k(\boldsymbol{x}_{\mathcal{A}, k}; \boldsymbol{y}_{\mathcal{A}, k}) = \mathbb{E}_{\boldsymbol{y}_{\mathcal{A}, k}}\left[\mathcal{D}_{\text{KL}}\left[p(\boldsymbol{x}_{\mathcal{A}, k}|\boldsymbol{y}_{\mathcal{A}, k})\|p(\boldsymbol{x}_{\mathcal{A}, k})\right]\right]\\
    = \mathbb{E}_{\boldsymbol{x}_{\mathcal{A}, k}}\left[\mathcal{D}_{\text{KL}}\left[p(\boldsymbol{y}_{\mathcal{A}, k}|\boldsymbol{x}_{\mathcal{A}, k})\|p(\boldsymbol{y}_{\mathcal{A}, k})\right]\right].\label{eqn:marginal_MI}
   \end{multlined}
\end{equation}
With this interpretation, we see that by maximizing the mutual information, we are effectively maximizing the expected dissimilarity between a prior and its conditional distribution. The intuition for its use as a reward function in our context lies in that the conditional density always contains that same or greater information content of the prior, so optimally discriminating the two provides the greatest information gain. 

Now, rather than considering a single measurement trial, we can apply this framework to all measurements collected over the entirety of an observation interval. Consider the sequence of augmented states and measurements collected during this period, and stacked into a vector form:
\begin{equation}
    \boldsymbol{X} = \left[\boldsymbol{x}_{\mathcal{A}, 1}^\top, \cdots, \boldsymbol{x}_{\mathcal{A}, N_{\text{meas}}}^\top \right]^\top,
\end{equation}
and
\begin{equation}
    \boldsymbol{Y} = \left[ \boldsymbol{y}_{\mathcal{A}, 1}^\top, \cdots, \boldsymbol{y}_{\mathcal{A}, N_{\text{meas}}}^\top \right],
\end{equation}
respectively. Under linearized Gaussian assumptions, the joint distribution 
\begin{equation}
    p\left(\boldsymbol{X}, \boldsymbol{Y} \right) = \mathcal{N}\left(\begin{bmatrix}
        \boldsymbol{X}\\
        \boldsymbol{Y}
    \end{bmatrix};
    \begin{bmatrix}
        \bar{\boldsymbol{X}}\\
        \bar{\boldsymbol{Y}}
    \end{bmatrix},
    \Tilde{\Sigma}
    \right),
\end{equation}
where,
\begin{equation}
    \Tilde{\Sigma} = \begin{bmatrix}
        \Tilde{P} & \Tilde{\Gamma}\\
        \Tilde{\Gamma}^\top & \Tilde{S}.
    \end{bmatrix} = 
    \begin{bmatrix}
        \Tilde{\Phi}\Tilde{P}\Tilde{\Phi}^{\top} & \Tilde{\Phi} \Tilde{P} \Tilde{H}^{\top}\\
        \Tilde{H}\Tilde{P}\Tilde{\Phi}^{\top} & \Tilde{H}\Tilde{P}\Tilde{H}^{\top} + \Tilde{R}
    \end{bmatrix}
\end{equation}
With this, the corresponding prior and conditional measurement densities are 
\begin{flalign}
    p\left(\boldsymbol{Y}\right) &= \mathcal{N}\left(\boldsymbol{Y}; \bar{\boldsymbol{Y}}, \Tilde{S} \right) \label{eqn:predicted_augmented_measurement} \\
    &= \mathcal{N}\left(\boldsymbol{Y}; \bar{\boldsymbol{Y}}, \Tilde{H} \Tilde{P} \Tilde{H}^\top + \Tilde{R} \right),\nonumber
\end{flalign}
and
\begin{flalign}
    p\left(\boldsymbol{Y}|\boldsymbol{X}\right) &= \mathcal{N}\left(\boldsymbol{Y}; \hat{\boldsymbol{Y}}, \Tilde{S} - \Tilde{\Gamma}^{\top} \Tilde{P}^{-1} \Tilde{\Gamma} \right)\\
    &= \mathcal{N}\left(\boldsymbol{Y}; \hat{\boldsymbol{Y}}, \Tilde{R} \right),\nonumber
\end{flalign}
respectively. Applying the form contained in the second line of Eq.~\ref{eqn:marginal_MI}, the mutual information between the stacked state vector and stacked measurement vector is 
\begin{equation}
   \begin{multlined}[b]
    \mathcal{I}(\boldsymbol{X}; \boldsymbol{Y}) = \mathbb{E}_{\boldsymbol{X}}\left[\mathcal{D}_{\text{KL}}\left[p(\boldsymbol{Y}|\boldsymbol{X})\|p(\boldsymbol{Y})\right]\right] =
    \frac{1}{2} \ln\left( \frac{|\Tilde{H} \Tilde{P} \Tilde{H}^{\top} + \Tilde{R}|}{|\Tilde{R}|}\right).\label{eqn:matrix_MI}
   \end{multlined}
\end{equation}
The block state error covariance matrix in Eq.~\ref{eqn:predicted_augmented_measurement} is defined as 
\begin{equation}
    \Tilde{P} = \text{diag}\left(P_{\mathcal{A}, 0}, Q_{\mathcal{A}, 1}, Q_{\mathcal{A}, 2} \cdots, Q_{\mathcal{A}, N_{\text{meas}}} \right),
\end{equation}
where the first block corresponds to the initial state error covariance of the sensor and targets: $ P_{\mathcal{A}, 0} = \text{diag}\left(P_{\mathcal{S}, 0}, P_{\mathcal{T}, 0}^{(1)},\cdots, P_{\mathcal{T}, 0}^{(N_{\mathcal{T}})}  \right)$. The subsequent diagonal blocks add the contributions attributed to the process noise: $Q_{\mathcal{A}, k} = \text{diag}\left(Q_{\mathcal{S}, k}, Q_{\mathcal{T}, k}^{(1)},\cdots, Q_{\mathcal{T}, k}^{(N_{\mathcal{T}})} \right)$. The block matrix providing partial derivatives of the stacked measurement with respect to the stacked state vector is defined as
\begin{equation}
    \Tilde{H} = \begin{bmatrix}
        H_{\mathcal{A}, 0} & \boldsymbol{0} & \boldsymbol{0} & \cdots & \boldsymbol{0}\\
        H_{\mathcal{A}, 1} \Phi_{\mathcal{A}}(t_1, t_0) & H_{\mathcal{A}, 1} & \boldsymbol{0} & \cdots & \boldsymbol{0}\\
        H_{\mathcal{A}, 2} \Phi_{\mathcal{A}}(t_2, t_0) & H_{\mathcal{A}, 2} \Phi_{\mathcal{A}}(t_2, t_1) & H_{\mathcal{A}, 2} & \cdots & \boldsymbol{0}\\
        \vdots & \vdots & \vdots & \ddots & \vdots\\
        H_{\mathcal{A}, N_{\text{meas}}}\Phi_{\mathcal{A}}(t_{N_{\text{meas}}}, t_0) & H_{\mathcal{A}, N_{\text{meas}}} \Phi_{\mathcal{A}}(t_{N_\text{meas}}, t_1) & H_{\mathcal{A}, N_{\text{meas}}} \Phi_{\mathcal{A}}(t_{N_\text{meas}}, t_2) & \cdots & H_{\mathcal{A}, N_{\text{meas}}}
    \end{bmatrix}\mbox{,}
    \label{eqn:big_H}
\end{equation}
where
\begin{equation}
    H_{\mathcal{A}, k} = \begin{bmatrix}
        \frac{\partial \boldsymbol{y}_k^{(1)}}{\partial \boldsymbol{x}_{\mathcal{S}, k}} & \frac{\partial \boldsymbol{y}_k^{(1)}}{\partial \boldsymbol{x}^{(1)}_{\mathcal{T}, k}} & \boldsymbol{0} & \cdots & \boldsymbol{0}\\
        \frac{\partial \boldsymbol{y}_k^{(2)}}{\partial \boldsymbol{x}_{\mathcal{S}, k}} & \boldsymbol{0} & \frac{\partial \boldsymbol{y}_k^{(2)}}{\partial \boldsymbol{x}^{(2)}_{\mathcal{T}, k}} &  \cdots & \boldsymbol{0}\\
        \vdots & \vdots & \vdots & \ddots & \vdots\\
        \frac{\partial \boldsymbol{y}_k^{(N_\mathcal{T})}}{\partial \boldsymbol{x}_{\mathcal{S}, k}} & \boldsymbol{0} & \boldsymbol{0} &  \cdots & \frac{\partial \boldsymbol{y}_k^{(N_\mathcal{T})}}{\partial \boldsymbol{x}^{(N_\mathcal{T})}_{\mathcal{T}, k}}\\
    \end{bmatrix},
\end{equation}
and
\begin{equation}
    \Phi_{\mathcal{A}}(t_k, t_\ell) = \text{diag}\left(\frac{\partial \boldsymbol{x}_{\mathcal{S}, k}}{\partial \boldsymbol{x}_{\mathcal{S}, \ell}}, \frac{\partial \boldsymbol{x}^{(1)}_{\mathcal{T}, k}}{\partial \boldsymbol{x}^{(1)}_{\mathcal{T}, \ell}}, \frac{\partial \boldsymbol{x}^{(2)}_{\mathcal{T}, k}}{\partial \boldsymbol{x}^{(2)}_{\mathcal{T}, \ell}}, \cdots, \frac{\partial \boldsymbol{x}^{(N_\mathcal{T})}_{\mathcal{T}, k}}{\partial \boldsymbol{x}^{(N_\mathcal{T})}_{\mathcal{T}, \ell}}  \right).
\end{equation}
Finally, the block measurement noise covariance
\begin{equation}
    \Tilde{R} = \text{diag}(R_{\mathcal{A}, 1}, R_{\mathcal{A}, 2}, \cdots, R_{\mathcal{A}, N_{\text{meas}}}),
    \label{eqn:big_R}
\end{equation}
where $R_{{\mathcal{A}}, k} = \text{diag}\left(R, R, \cdots, R\right)$, containing the measurement noise covariance for a single measurement, repeated $N_\mathcal{T}$ times.


%% file: convex_programming.tex

\subsection{Linearization}

This section describes an approach for transforming the continuous time optimal control problem presented in Eq.~\ref{eqn:general_noncovex_continuous_problem} into a sequence of convex, static parameter optimization problems. We use the \ac{SCvx} algorithm; an \ac{SCP} approach with global convergence and superlinear convergence rate \cite{Mao_2018}. For this discussion, we follow the notation presented by the authors in Reference \citenum{Malyuta_2022} and refer the reader to their paper as a helpful guide for implementing \ac{SCvx}. To start, first consider the linearized approximation of our original optimal control problem: 
\begin{subequations}
    \begin{flalign}
        &\hspace{25ex} \min_{\boldsymbol{u}(t)} J (\boldsymbol{x}(t), \boldsymbol{u}(t))&\\
        &\hspace{26ex} \mbox{s.t.} \hspace{1ex} \dot{\boldsymbol{x}}(t) = A(t)\boldsymbol{x}(t) + B(t)\boldsymbol{u}(t) + \boldsymbol{r}(t) \mbox{,}&\\
        &\hspace{30ex} \boldsymbol{g}_{\text{ic}}(\boldsymbol{x}(t_0)) = \boldsymbol{0}\mbox{,}&\\
        &\hspace{30ex} \boldsymbol{g}_{\text{tc}}(\boldsymbol{x}(t_f)) = \boldsymbol{0}\mbox{,}&\\
        &\hspace{30ex} \|\boldsymbol{u}(t)\|_2 \leq a_{\text{max}}\mbox{,} &\\
        &\hspace{30ex} \|\delta \boldsymbol{x}(t)\|_2 + \|\delta \boldsymbol{u}(t)\|_2 \leq \eta\mbox{.}\label{eqn:continous_time_step_contraint}&
    \end{flalign}
    \label{eqn:linearized_continuous_form}%
\end{subequations}
In the above, we drop the subscript denoting the sensor and targets for compactness while assuming that the state and control variables in this section correspond to the controllable sensor platform. The linearization variables are defined as 
\begin{subequations}
    \begin{flalign}
        A(t) &= \frac{\partial \boldsymbol{f}(t, \boldsymbol{x}(t), \boldsymbol{u}(t))}{\partial \boldsymbol{x}(t)} \bigg|_{\bar{\boldsymbol{x}}(t), \bar{\boldsymbol{u}}(t)}\mbox{,}\\
        B(t) &= \frac{\partial \boldsymbol{f}(t, \boldsymbol{x}(t), \boldsymbol{u}(t))}{\partial \boldsymbol{u}(t)}\bigg|_{\bar{\boldsymbol{x}}(t), \bar{\boldsymbol{u}}(t)} \mbox{,}\\
        \boldsymbol{r}(t) &= \boldsymbol{f}(t, \bar{\boldsymbol{x}}(t), \bar{\boldsymbol{u}}(t)) - A\bar{\boldsymbol{x}}(t) - B\bar{\boldsymbol{u}}(t)\mbox{.}
    \end{flalign}
    \label{eqn:linearization_definitions}%
\end{subequations}
Linearization accuracy is only valid in a local neighborhood of the reference state and control variables, $\bar{\boldsymbol{x}}(t)$ and $\bar{\boldsymbol{u}}(t)$. Therefore, \ac{SCvx} imposes a trust region with Eq.~\ref{eqn:continous_time_step_contraint}, where 
\begin{subequations}
    \begin{flalign}
        \delta \boldsymbol{x}(t) &= \boldsymbol{x}(t) - \bar{\boldsymbol{x}}(t)\mbox{,}\\
        \delta \boldsymbol{u}(t) &= \boldsymbol{u}(t) - \bar{\boldsymbol{u}}(t)\mbox{.}
    \end{flalign}
    \label{eqn:delta_states_and controls}%
\end{subequations}
The parameter $\eta$ is recursively adjusted at each convex step based on a trust-region update described later.


\subsection{Discretization}

We use a first-order hold interpolation scheme to approximate the continuous problem in Eq.~\ref{eqn:linearized_continuous_form} at discrete time nodes $\{t_k\}_{k = 1}^N$ so that, between nodes $k$ and $k + 1$, the continuous time control and dynamics are
\begin{subequations}
    \begin{flalign}
        \boldsymbol{u}(t) &= \frac{t_{k + 1} - t}{t_{k + 1} - t_k}\boldsymbol{u}_k + \frac{t - t_k}{t_{k + 1} - t_k}\boldsymbol{u}_{k + 1} \nonumber \\ 
        &= \lambda_k^{-}(t)\boldsymbol{u}_k + \lambda_k^{+}(t)\boldsymbol{u}_{k + 1}\mbox{,}\\
        \dot{\boldsymbol{x}}(t) &= A(t)\boldsymbol{x}(t) + B(t)\lambda_k^{-}(t)\boldsymbol{u}_k 
                    + B(t)\lambda_k^{+}\boldsymbol{u}_{k + 1} + \boldsymbol{r}(t)\mbox{.}\label{eqn:first_order_hold_continuous_time}
    \end{flalign}
\label{eqn:first_order_control_interp}%
\end{subequations}
Provided this, the discrete time state transition is 
\begin{equation}
    \boldsymbol{x}_{k + 1} = A_k \boldsymbol{x}_k + B_k^{-} \boldsymbol{u}_k + B_k^{+} \boldsymbol{u}_{k + 1} + \boldsymbol{r}_k \mbox{,}\\
\end{equation}
where
\begin{subequations}
    \begin{flalign}
         A_k &= \Phi(t_{k + 1}, t_k)\mbox{,}\\
         B_k^{-} &= A_k\int_{t_k}^{t_{k + 1}} \Phi(\tau, t_k)^{-1}
         B(\tau)\lambda_{k}^-(\tau) d\tau \mbox{,}\\
         B_k^{+} &= A_k\int_{t_k}^{t_{k + 1}} \Phi(\tau, t_k)^{-1} B(\tau) \lambda_{k}^{+}(\tau) d\tau\mbox{,} \\
         \boldsymbol{r}_k &= A_k \int_{t_k}^{t_{k + 1}} \Phi(\tau, t_k)^{-1} \boldsymbol{r}(\tau) d\tau.
    \end{flalign}
    \label{eqn:discrete_integral_definitions}%
\end{subequations}
One common feature of \ac{SCP} algorithms is the phenomenon known as \textit{artificial infeasibility}. Under this condition, the linearized path constraints may not admit a feasible solution, despite the feasibility of the original control problem. \ac{SCvx} makes the algorithmic choice to augment the discrete state transition with a slack variable, referred to in the literature as a “virtual control”, which we denote as $\epsilon_k$. The virtual control serves to soften the constraint, and in effect, allows feasibility at intermediate convex steps. Consequently, the final form of the local convex optimization problem is
\begin{subequations}
    \begin{flalign}
        &\hspace{20ex} \min_{\boldsymbol{x}, \boldsymbol{u}, \boldsymbol{\epsilon} } \mathcal{L}(\boldsymbol{x}, \boldsymbol{u}, \boldsymbol{\epsilon})&\\
        &\hspace{21ex}\mbox{s.t.} \hspace{1ex} \boldsymbol{x}_{k + 1} = A_k \boldsymbol{x}_k + B_k^{-}\boldsymbol{u}_k + B_k^{+} \boldsymbol{u}_{k + 1} + \boldsymbol{r}_k + E_k \boldsymbol{\epsilon}_k\mbox{,}&\\
        &\hspace{25ex} \boldsymbol{g}_{\text{ic}}(\boldsymbol{x}_0) = \boldsymbol{0}\mbox{,}&\\
        &\hspace{25ex} \boldsymbol{g}_{\text{tc}}(\boldsymbol{x}_f) = \boldsymbol{0}\mbox{,}&\\
        &\hspace{25ex} \|\boldsymbol{u}_k\|_2 \leq u_{max, k}\mbox{,} &\label{eqn:thrust_constraint_in_linear_objective}\\
        &\hspace{25ex} \|\delta \boldsymbol{x}_k \|_2 + \| \delta \boldsymbol{u}_k \|_2 \leq \eta \mbox{.} &
    \end{flalign}
    \label{eqn:general_convex_problem}%
\end{subequations}
Here, the matrix $E_k$ is the virtual control gain matrix, which we compute as 
\begin{equation}
    E_k = A_k \int_{t_k}^{t_{k+1}} \Phi(\tau, t_k)^{-1}E(\tau) d\tau \mbox{,}\label{eqn:virtual_control_gain}%
\end{equation}
and $E(t)$ is the identity matrix by convention. The function $\mathcal{L}(\cdot)$ is the linearized discrete time objective function, further detailed in a later section. The variables $\boldsymbol{x}$, $\boldsymbol{u}$, and $\boldsymbol{\epsilon}$ represent the augmented discrete states, control inputs, and virtual controls along the trajectory arc, respectively.

In practice, we jointly compute the discrete time variables in Eq.~\ref{eqn:discrete_integral_definitions} and Eq.~\ref{eqn:virtual_control_gain} at each convex step by numerically integrating the system of ordinary differential equations,
\begin{equation}
    \dot{\boldsymbol{\Xi}}(t) = \begin{bmatrix}
        \boldsymbol{f}(\boldsymbol{x}(t), \bar{\boldsymbol{u}}(t))\\
        \text{flat}\big(A(t)\Phi(t, t_k)\big)\\
        \text{flat}\big(\Phi(t, t_k)^{-1}\lambda_k^{-}(t) B(t)\big)\\
        \text{flat}\big(\Phi(t, t_k)^{-1}\lambda_k^{+}(t) B(t)\big)\\
        \Phi(t, t_k)^{-1}\boldsymbol{r}(t)\\
        \text{flat}\left(\Phi(t, t_k)^{-1} E(t)\right),\label{eqn:matrix_differential_eqn}
    \end{bmatrix},
\end{equation}
with initial conditions
\begin{equation}
    \boldsymbol{\Xi}(t_k) = \begin{bmatrix}
        \bar{\boldsymbol{x}}_k\\
        \text{flat}(I_{n_x\times n_x})\\
        \boldsymbol{0}_{n_x n_u \times 1}\\
        \boldsymbol{0}_{n_x n_u \times 1}\\
        \boldsymbol{0}_{n_x \times 1}\\
        \boldsymbol{0}_{n_x n_x\times 1}
    \end{bmatrix}\mbox{.}
\end{equation}
Here, the operator $\text{flat}(\cdot)$ stacks the columns of a matrix. The integration is repeated between each successive interval spanning the discrete time set, $\{t_k\}_{k = 1}^N$. A computational challenge in computing the terms in Eq.~\ref{eqn:matrix_differential_eqn} arises from the matrix inversion of the state transition matrix. This task can be efficiently managed using a LU decomposition and Gaussian elimination \cite{Reynolds_2020}.


\subsection{Trust Region Update}

As noted previously, the trust region radius, $\eta$, bounds the change in the decision variables at each convex step. Here, we detail the algorithmic procedure used by \ac{SCvx} for selecting $\eta$ based on the accuracy of the linearized approximation. A linearization defect is defined as 
\begin{equation}
    \boldsymbol{\delta}_k = \boldsymbol{x}_{k + 1} - \boldsymbol{f}_k(\boldsymbol{x}_k, \boldsymbol{u}_k, \boldsymbol{u}_{k+ 1})\mbox{,}
\end{equation}
where $\boldsymbol{f}_k(\cdot)$ is the true (not linearized) discrete time transition between nodes $k$ and $k + 1$,  defined as
\begin{equation}
    \boldsymbol{f}_k(\boldsymbol{x}_k, \boldsymbol{u}_k, \boldsymbol{u}_{k+ 1}) = \int_{t_k}^{t_{k + 1}} \boldsymbol{f}(\boldsymbol{x}(\tau), \boldsymbol{u}(\tau), \tau) d\tau, 
\end{equation}
and $\boldsymbol{u}(\cdot)$ is given by the first-order hold interpolation in Eq.~\ref{eqn:first_order_control_interp}. The expected cost improvement, produced by the convex program step, is compared with the actual improvement through the linearization accuracy metric,
\begin{equation}
    \rho =\frac{\mathcal{J}(\bar{\boldsymbol{x}}, \bar{\boldsymbol{u}}, \bar{\boldsymbol{\delta}}) - \mathcal{J}(\boldsymbol{x}^*, \boldsymbol{u}^*, \boldsymbol{\delta}^*)}{\mathcal{J}(\bar{\boldsymbol{x}}, \bar{\boldsymbol{u}}, \bar{\boldsymbol{\delta}}) - \mathcal{L}(\boldsymbol{x}^{*}, \boldsymbol{u}^{*}, \boldsymbol{\epsilon}^*)}\mbox{.}\label{eqn:accuracy_metric}
\end{equation}
The discretized true nonlinear cost $\mathcal{J}(\cdot)$ is detailed in the subsequent section. Variables marked with the superscript $(*)$ represent those generated by each iteration of the convex solver. Based on the accuracy metric, we adjust the reference states and trust region accordingly, as outlined in Table~\ref{tab:trust_region_update_rules}. This is consistent with the guidelines in Reference \citenum{Malyuta_2022}. The parameters $\rho_0$, $\rho_1$, and $\rho_2$ are user-defined variables within the range $[0, 1]$, while $\beta_{\text{sh}}$ and $\beta_{\text{gr}}$ denote the trust region shrinkage and growth factors, respectively.

\begin{table}[h]
\caption{\label{tab:trust_region_update_rules} Trust region update protocol}
\centering
\begin{tabular}{lcccccc}
\hline
Case & &  & Update Rule &\\\hline
$\rho < \rho_0$ && $\eta \rightarrow \eta/\beta_{\text{sh}}$ & $\bar{\boldsymbol{x}} \rightarrow \bar{\boldsymbol{x}}$ & $\bar{\boldsymbol{u}} \rightarrow \bar{\boldsymbol{u}}$\\
$\rho_0 \leq \rho < \rho_1$ && $\eta \rightarrow \eta/\beta_{\text{sh}}$ & $\bar{\boldsymbol{x}} \rightarrow \boldsymbol{x}^*$ & $\bar{\boldsymbol{u}} \rightarrow \boldsymbol{u}^*$\\
$\rho_1 \leq \rho < \rho_2$ && $\eta \rightarrow \eta$ & $\bar{\boldsymbol{x}} \rightarrow \boldsymbol{x}^*$ & $\bar{\boldsymbol{u}} \rightarrow \boldsymbol{u}^*$\\
$\rho_2 \leq \rho$ && $\eta \rightarrow \beta_{\text{gr}} \eta$ & $\bar{\boldsymbol{x}} \rightarrow \boldsymbol{x}^*$ & $\bar{\boldsymbol{u}} \rightarrow \boldsymbol{u}^*$\\
\hline
\end{tabular}
\end{table}

\subsection{Selection of Discrete Time Nodes}

We use a physically-informed approach for spacing the discrete time nodes based on the generalized Sundman transformation \cite{Szebehely_1969}. In highly nonlinear portions of the observer's trajectory, the linearization approximations used by \ac{SCvx} do not adequately approximate the true dynamics when nodes are sparse. The Sundman transformation acts as a simple, yet effective, approach for step-size regulation by increasing node resolution in those nonlinear regions. The temporal regularization is carried out through the transformation 
\begin{equation}
    dt = r_m^\sigma d\tau \mbox{,}
\end{equation}
where $r_m$ is the distance of the observer from the Moon, and $\sigma$ is a user-defined scalar. There exists a one-to-one equivalence between $t$ and $\tau$ when $\sigma = 0$, however, a positive $\sigma$ acts to dilate the time variable near perilune -- the portion of the trajectory where we expect the greatest nonlinearity. The time-node spacing is computed by numerically integrating the ordinary differential equation,
\begin{equation}
    \frac{d \boldsymbol{z}}{d \tau} = r^\alpha_m \begin{bmatrix}
        \boldsymbol{f}(\boldsymbol{x}(\tau), \tau)\\
        1
    \end{bmatrix}\mbox{,}%
\end{equation}
where,
\begin{equation}
    \boldsymbol{z}(\tau) = [\boldsymbol{x}^\top(\tau), t]^\top\mbox{.}%
\end{equation}
%
%

The Sundman transformation results in non-uniform time intervals between consecutive nodes. Therefore, to correctly model the thrust constraints in Eq.~\ref{eqn:thrust_constraint_in_linear_objective}, we must determine $u_{\text{max}, k}$ such that it aligns with the achievable impulse between collocation points. Using a trapezoidal approximation, this can be written as
\begin{equation}
    \frac{u_{\text{max}, k} + u_{\text{max}, k + 1}}{2 \Delta t_k} = a_{\text{max}}{,}
\end{equation}
and can be solved at all nodes with, 
\begin{equation}
    \begin{bmatrix}
        \sfrac{1}{\Delta t_1} & \sfrac{1}{\Delta t_1} & 0 & \cdots & 0 & 0\\
        0 & \sfrac{1}{\Delta t_2} & \sfrac{1}{\Delta t_2} & \cdots & 0 & 0\\
        \vdots & \vdots & \ddots  & \ddots  & \vdots  & \vdots\\
        0 & 0 & \cdots & 0 & \sfrac{1}{\Delta t_{N - 1}} & \sfrac{1}{\Delta t_{N - 1}} 
    \end{bmatrix}
    \begin{bmatrix}
        u_{\text{max}, 1}\\
        u_{\text{max}, 2}\\
        \vdots\\
        u_{\text{max}, N}
    \end{bmatrix} = 2 a_{\text{max}}
    \begin{bmatrix}
         1\\
         1\\
         \vdots\\
         1
    \end{bmatrix}\mbox{.}\label{eqn:linear_system_for_thrust_bounds}
\end{equation}
Eq. \ref{eqn:linear_system_for_thrust_bounds} is an underdetermined linear system of equations, and its minimum norm solution can be found with a QR factorization. 

%% file: convex_objective.tex
This section describes the true nonlinear objective and its corresponding linearized approximation, both used in the previous section. As mentioned, we use a convex combination of the control effort exercised, and the information gain generated through a sequence of observations. Additionally, for the final solution to be dynamically feasible, the nonlinear defects, $\delta$, must be driven to zero. With these design considerations, the true discretized \ac{SCvx} nonlinear objective is
\begin{flalign}
    \mathcal{J}(\boldsymbol{x}_{\mathcal{S}}, \boldsymbol{u}_{\mathcal{S}}, \boldsymbol{\delta}) = -\alpha_h \mathcal{I}(\boldsymbol{X}; \boldsymbol{Y}) +
    &(1 - \alpha_h)\sum_{k = 1}^{N - 1}\frac{\Delta t_k}{2} \left(||\boldsymbol{u}_{\mathcal{S}, k}||_2 + ||\boldsymbol{u}_{\mathcal{S}, k + 1}||_2\right) +\label{eqn:scvx_nonlinear_cost}\\ 
    &\gamma \sum_{k = 1}^{N - 1}\frac{\Delta t_k}{2} \left(||\boldsymbol{\delta}_k||_1 + ||\boldsymbol{\delta}_{k + 1}||_1\right).\nonumber
\end{flalign}
Here, we use a trapezoid integration to determine the contributions of the total thrust impulse and the defect cost in the objective function. Note the negative sign distributed to the information gain term, consistent with the minimization of $\mathcal{J}(\cdot)$. The parameter $\gamma$ is a positive scalar chosen to be sufficiently large to drive the defects to zero. 

Each \ac{SCvx} generates a first-order approximation of Eq.~\ref{eqn:scvx_nonlinear_cost} for each convex solver iteration. A linearized approximation around the previous guess is provided by 
\begin{flalign}
    \mathcal{L}(\boldsymbol{x}_{\mathcal{S}}, \boldsymbol{u}_{\mathcal{S}}, \boldsymbol{\epsilon}) = -\alpha_h \mathcal{I}_{\text{approx}}(\boldsymbol{X}; \boldsymbol{Y}) +
    &(1 - \alpha_h)\sum_{k = 1}^{N - 1}\frac{\Delta t_k}{2} \left(||\boldsymbol{u}_{\mathcal{S}, k}||_2 + ||\boldsymbol{u}_{\mathcal{S}, k + 1}||_2\right) +\label{eqn:linearized_objective}\\ 
    &\gamma \sum_{k = 1}^{N - 1}\frac{\Delta t_k}{2} \left(||E_k \boldsymbol{\epsilon}_k||_1 + ||E_{k + 1} \boldsymbol{\epsilon}_{k + 1}||_1\right).\nonumber
\end{flalign}

The primary challenge that we address in the following is deriving a first-order approximation of the mutual information of the observer, $\mathcal{I}_{\text{approx}}(\cdot)$. While not required, for simplicity, we assume that over the course of an observation period the observer is passive (i.e., no thrust input). Therefore, the observer states at each subsequent measurement time are determined by solely the state of the observer at the starting epoch of the observation period, $\boldsymbol{x}_{\mathcal{S}, k^*}$. The first-order mutual information approximation is then provided by
\begin{equation}
    \mathcal{I}_{\text{approx}}(\boldsymbol{X}; \boldsymbol{Y}) = \mathcal{I}(\boldsymbol{X}, \boldsymbol{Y})\bigg|_{\bar{\boldsymbol{x}}_{\mathcal{S}, k^*}} + 
    \frac{\partial \mathcal{I}(\boldsymbol{X}; \boldsymbol{Y})}{\partial \boldsymbol{x}_{\mathcal{S}, k^*}}\bigg|_{\bar{\boldsymbol{x}}_{\mathcal{S}, k^*}} \delta \boldsymbol{x}_{\mathcal{S}, k^*},
\end{equation}
where, using Eq.~\ref{eqn:matrix_MI}, 
\begin{equation}
    \frac{\partial \mathcal{I}(\boldsymbol{X}; \boldsymbol{Y})}{\partial \boldsymbol{x}_{\mathcal{S}, k^*}} = \frac{\partial}{\partial \boldsymbol{x}_{\mathcal{S}, k^*}}\left[\frac{1}{2}\left(\ln|\Tilde{H}\Tilde{P}\Tilde{H}| - \ln|\Tilde{R}|  \right)\right].\label{eqn:MI_gradient}
\end{equation}
For a scalar value $\xi$, and some matrix $M$ that is a function of this scalar, the matrix determinant derivative \cite{Petersen_2008}
\begin{equation}
    \frac{d}{d \xi} \mathrm{det}(M(\xi)) = \mathrm{det}(M(\xi))\cdot \mathrm{tr}\left(M^{-1}(\xi) \frac{d}{d \xi} M(\xi)\right).
\end{equation}
Using this identity by extension, the gradient in Eq.~\ref{eqn:MI_gradient} is 
\begin{equation}
    \frac{\partial \mathcal{I}(\boldsymbol{X}; \boldsymbol{Y})}{\partial \boldsymbol{x}_{\mathcal{S}, k^*}}  = \frac{1}{2} \left[\sum_{j = 1}^m T_{j, j, 1}, \cdots, \sum_{j = 1}^m T_{j, j, n} \right].
\end{equation}
The tensor array $T \in \mathbb{R}^{m\times m\times n_x}$ is
\begin{equation}
    T = \left(\Tilde{H}\Tilde{P}\Tilde{H}^{\top} + \Tilde{R}\right)^{-1}\times_1 \left[\frac{\partial \Tilde{H}}{\partial \boldsymbol{x}_{\mathcal{S}, k^*}} \times_1 \Tilde{P} \Tilde{H}^\top + \Tilde{H} \Tilde{P} \times_1 \frac{\partial \Tilde{H}^\top}{\partial \boldsymbol{x}_{\mathcal{S}, k^*}} \right].
\end{equation}
The operation $B\times_1 C$ denotes the mode-1 tensor/matrix product. Computation of the tensor $\frac{\partial \Tilde{H}}{\partial \boldsymbol{x}_{\mathcal{S}, k^*}}$ is rather involved. Consider the second row, first column matrix block 
\begin{equation}
    \Tilde{H}^{2, 1} = H_{\mathcal{A}, 1}\Phi_{\mathcal{A}}(t_1, t_*) = H_{\mathcal{A}, 1} \frac{\partial \boldsymbol{x}_{\mathcal{A}, 1}}{\partial \boldsymbol{x}_{\mathcal{A}, k^*}}
\end{equation}
from Eq.~\ref{eqn:big_H}. Then its corresponding partial derivative is
\begin{equation}
    \frac{\partial \Tilde{H}^{2, 1}}{\partial \boldsymbol{x}_{\mathcal{S}, k^*}} = \left(\frac{\partial H_{\mathcal{A}, 1}}{\partial \boldsymbol{x}_{\mathcal{A}, 1}} \times_2 
    \frac{\partial \boldsymbol{x}_{\mathcal{A}, 1}}{\partial \boldsymbol{x}_{\mathcal{S}, k^*}}\right) \times_1 \frac{\partial \boldsymbol{x}_{\mathcal{A}, 1}}{\partial \boldsymbol{x}_{\mathcal{A}, k^*}} +
    H_{\mathcal{A}, 1} \times_1 \frac{\partial^2\boldsymbol{x}_{\mathcal{A}, 1}}{\partial \boldsymbol{x}_{\mathcal{A}, k^*} \partial \boldsymbol{x}_{\mathcal{S}, k^*}}
\end{equation}
where
\begin{equation}
    \frac{\partial \boldsymbol{x}_{\mathcal{A}, 1}}{\partial \boldsymbol{x}_{\mathcal{S}, k^*}} = \begin{bmatrix}
        \frac{\partial \boldsymbol{x}_{\mathcal{S}, 1}}{\partial \boldsymbol{x}_{\mathcal{S}, k^*}}\\
        \boldsymbol{0}\\
        \vdots \\
        \boldsymbol{0}
    \end{bmatrix}
\end{equation}
and, 
\begin{equation}
    \frac{\partial^2\boldsymbol{x}_{\mathcal{A}, 1}}{\partial \boldsymbol{x}_{\mathcal{A}, k^*} \partial \boldsymbol{x}_{\mathcal{S}, k^*}} = \begin{bmatrix}
        \begin{bmatrix}
            \frac{\partial^2 \boldsymbol{x}_{\mathcal{S}, 1}}{\partial \boldsymbol{x}^2_{\mathcal{S}, k^*}} & \boldsymbol{0} & \cdots & \boldsymbol{0}\\
            \boldsymbol{0} & \boldsymbol{0} & \cdots & \boldsymbol{0}\\
            \vdots & \vdots & \ddots & \vdots\\
            \boldsymbol{0} & \boldsymbol{0} & \cdots & \boldsymbol{0}
        \end{bmatrix}, &
        \begin{bmatrix}
            \boldsymbol{0}
        \end{bmatrix}, &
        \cdots, &
        \begin{bmatrix}
            \boldsymbol{0}
        \end{bmatrix}
    \end{bmatrix}.
\end{equation}
The matrix $\frac{\partial \boldsymbol{x}_{\mathcal{S}, 1}}{\partial \boldsymbol{x}_{\mathcal{S}, k^*}}$ and tensor $\frac{\partial^2 \boldsymbol{x}_{\mathcal{S}, 1}}{\partial \boldsymbol{x}^2_{\mathcal{S}, k^*}}$ are the first- and second-order state transition derivatives of the sensor platform state at the observation time with respect to the starting epoch state, respectively. The above procedure is repeated for all matrix blocks in Eq.~\ref{eqn:big_H}.  It should be noted that this formulation does not account for the sensitivity of the process noise covariance, implicitly associated through $\Tilde{P}$, with respect to $\boldsymbol{x}_{S, k^*}$.

%% file: numerical_cases.tex
\subsection{Equations of Motion}

We apply the methods introduced in our work to two test cases that consider spacecraft operating in the Earth-Moon three-body system. Our approach imposes no restrictions on the model fidelity, however, for simplicity, we choose to model the dynamical behavior of both the observer and targets using the \ac{CRTBP}. The \ac{CRTBP} assumes that the two primary bodies (i.e., the Earth and Moon) revolve in perfect circles around a common barycenter. The equations of motion of a third body are described in a rotating reference frame in which the $x$-axis points from the system barycenter to the Moon, the $z$-axis is oriented with the system's angular momentum vector, and the $y$-axis completes the right-hand. These are given by
\begin{subequations}
    \begin{flalign*}
        \ddot{x} &= 2\dot{y} + x - (1 - \mu)\frac{x + \mu}{r_1^3} - \mu \frac{x + \mu - 1}{r_2^3},\\
        \ddot{y} &= -2\dot{x} + y - (1 - \mu)\frac{y}{r_1^3} - \mu \frac{y}{r_2^3},\\
        \ddot{z} &= -(1 - \mu)\frac{z}{r_1^3} - \mu \frac{z}{r_2^3},
    \end{flalign*}
    \label{eqn:CRTBP_eom}%
\end{subequations}
where
\begin{subequations}
    \begin{flalign*}
        r_1^2 &= (x + \mu)^2 + y^2 + z^2,\\
        r_2^2 &= (x + \mu - 1)^2 + y^2 + z^2.
    \end{flalign*}
    \label{eqn:CRTBP_eom}%
\end{subequations}
The parameter $\mu = \frac{m_2}{m_1 + m_2}$, where $m_1$ is the mass of the Earth, and $m_2$ is the mass of the Moon. As is common practice in the astrodynamics community, we implement the model using normalized units.  The distance unit, $\text{DU}$, is defined by the distance of the Earth to the Moon, and the time unit is $\frac{1}{2\pi} P_{\text{synodic}}$, where $P_{\text{synodic}}$ is the period of the system around the barycenter. 

\subsection{Performance Evaluation}

To evaluate the effectiveness of the mutual information for trajectory planning, we perform a linear covariance analysis for a sequential estimator that considers the joint problem of simultaneously estimating the state of the observer and targets. In our case, this is equivalent to a \ac{CRLB} analysis, which, for nonlinear systems with Gaussian process and measurement noise, is generated through extended Kalman filter's covariance propagation and update equations, linearized around the true system state \cite{Taylor_1978}. The propagation and update are
\begin{equation*}
    \Bar{P}_{\mathcal{A}, k + 1} = \Phi_{A}(t_{k + 1}, t_k) \hat{P}_{\mathcal{A}, k} \Phi_{A}^{\top}(t_{k + 1}, t_k) + Q_{\mathcal{A}, k},
\end{equation*}
and,
\begin{equation*}
    \hat{P}_{\mathcal{A}, k} = (I - K_{\mathcal{A}, k} H_{{\mathcal{A}, k}}) \bar{P}_{\mathcal{A}, k},
\end{equation*}
respectively, where the Kalman gain is
\begin{equation*}
    K_{\mathcal{A}, k} = \bar{P}_{\mathcal{A}, k } H_{\mathcal{A}, k}^\top (H_{\mathcal{A}, k} \bar{P}_{\mathcal{A}, k } H_{\mathcal{A}, k} + R_{\mathcal{A}})^{-1}.
\end{equation*}
Additionally, we evaluate the total impulse of the observer executed over the scenario length. For this, we use the same trapezoid integration approximation provided in Eq.~\ref{eqn:scvx_nonlinear_cost}. 

\subsection{Test Case \#1}

The first test consists of an observer and one target spacecraft operating in adjacent \acp{DRO}, where the term ``adjacent" here refers to periodic orbits with similar values of the Jacobi integral and period. The observer operates over two periods of its initial reference orbit, and is slated to target a final state contained on a third \ac{DRO}. We use a single observation window in the time span $[3P_{\text{ref}}/4, 3P_{\text{ref}}/2]$, where $P_{\text{ref}}$ is the period of the initial reference orbit. In this window, we impose that $a_{\text{max}} = 0.0$. The relative position of the target serves as the measurement, which is sampled once per day. Such a sensing system may be composed of a monocular camera co-equipped with a laser ranging system. We assume that measurements have an expected \ac{RMS} error of $100$ m in each positional direction. Table~\ref{tab:test_case_1_orbits} summarizes the orbital properties for test case parameters. Table~\ref{tab:test_case_1_params} provides other relevant parameters for the test case. 

\begin{table}[h!]
\caption{\label{tab:test_case_1_orbits} Orbit Parameters for Test Case \#1}
\centering
\setlength{\tabcolsep}{2pt}
\begin{tabular}{ccccccc}
\hline
\hspace{1ex} & $x_0$ [DU] & $y_0$ [DU] &$z_0$ [DU] & $\dot{x}_0$ [DU/TU] &$\dot{y}_0$ [DU/TU] & $\dot{z}_0$ [DU/TU]  \\\hline
$\boldsymbol{x}_{\mathcal{S}}(t_0)$ & $7.78185828\times10^{-1}$ & 0.0 & 0.0 & 0.0 & $5.28996986\times10^{-1}$ & 0.0 \\
$\boldsymbol{x}_{\mathcal{S}}(t_f)$ &  $7.77831224\times10^{-1}$ & 0.0 & 0.0 & 0.0  &$5.56449590\times10^{-1}$ & 0.0\\
$\boldsymbol{x}_{\mathcal{T}}^{(1)}(t_0)$ & $7.78008526\times10^{-1}$ & 0.0 & 0.0 & 0.0 & $5.56190606\times10^{-1}$ & 0.0 \\
\hline
\end{tabular}
\end{table}

\begin{table}[h!]
\caption{\label{tab:test_case_1_params} Scenario Parameters for Test Case \#1}
\centering
\begin{tabular}{lcccccc}
\hline
Parameter & & & Symbol & Value\\\hline
Initial Position RMS (Sensor) &&& $1\sigma$ [km] & 100 \\
Initial Velocity RMS (Sensor) &&& $1\sigma$ [km/s] & $1\times10^{-2}$\\
Initial Position RMS (Target) &&& $1\sigma$ [km] & 100\\
Initial Velocity RMS (Target) &&& $1\sigma$ [km/s] & $1\times10^{-2}$\\
Maximum Thrust Accel. &&& $a_{\text{max}}$ [km/$s^2$] & $1\times10^{-6}$ \\
Process Noise Power Density &&& $Q$ [km/$s^3$]  & $1\times10^{-11}$\\
Measurement Noise RMS (x, y, z) &&& $1\sigma$ [m]  & 100 \\
Measurement Cadence &&& $f_{\text{meas}}$ [1/day] & 1.0 \\

\hline
\end{tabular}
\end{table}

\begin{figure}[hbt!]
\centering
\includegraphics[width=0.9\textwidth]{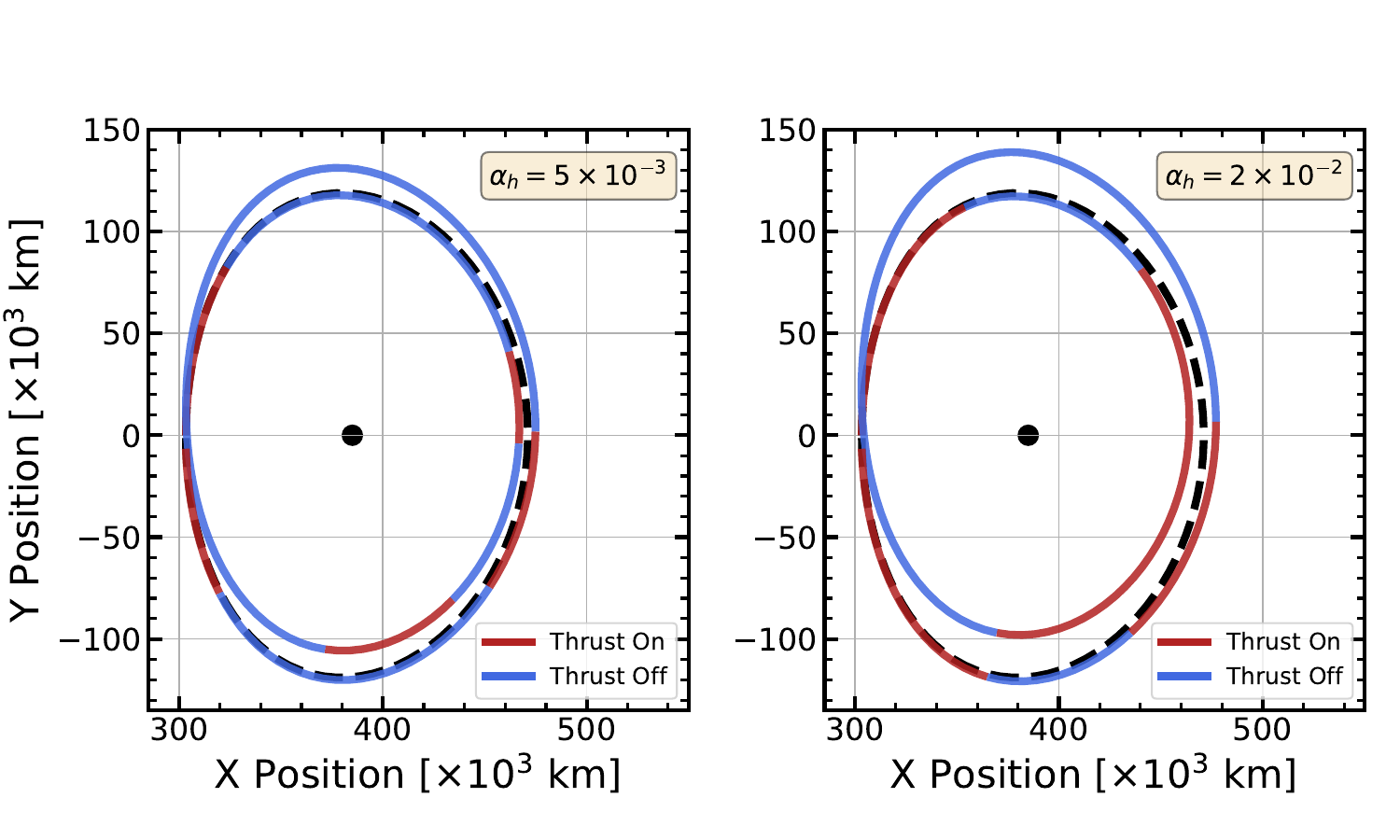}
\caption{Absolute trajectory of the observer for $\alpha_h = 5\times10^{-3}$ (left), and $\alpha_h = 2\times10^{-2} $ (right). The black dashed line plots the initial reference \ac{DRO}. }\label{fig:DRO_single_target_absolute_motion}
\end{figure}
\begin{figure}[hbt!]
\centering
\includegraphics[width=0.55\textwidth]{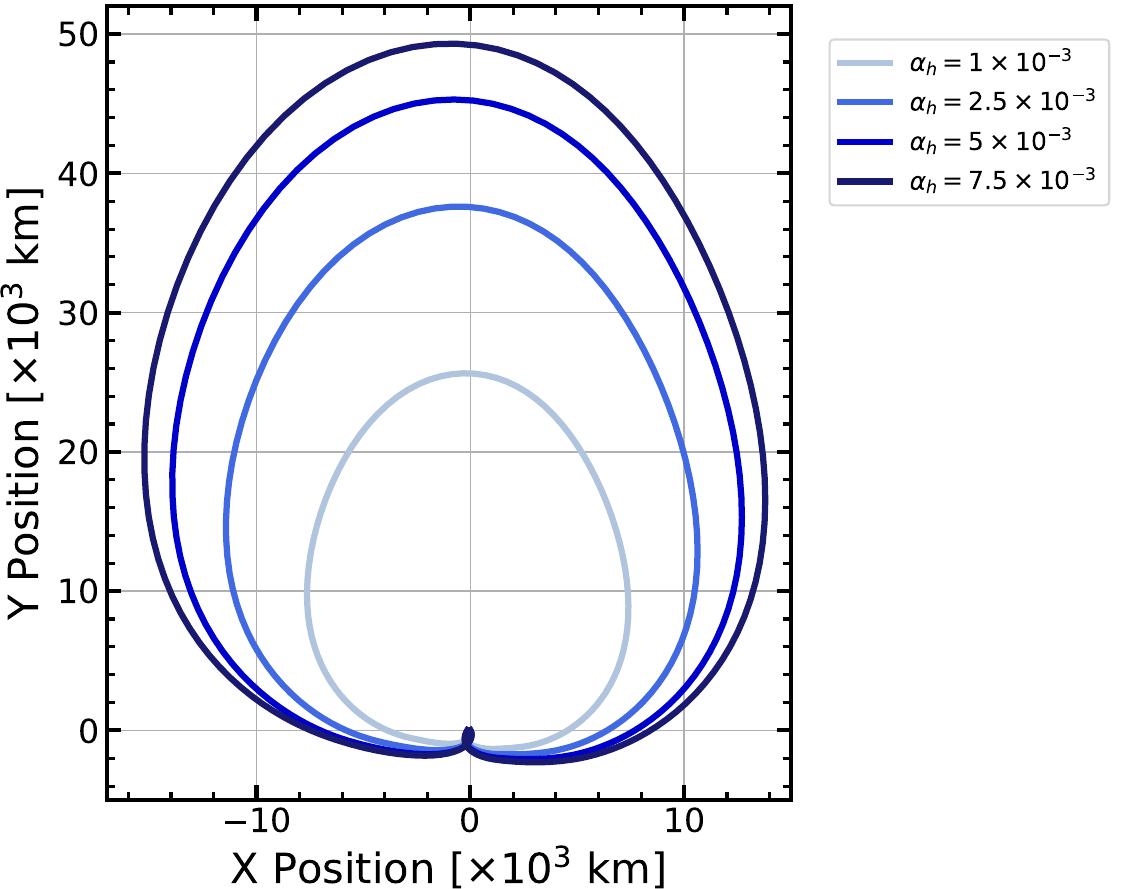}
\caption{Relative position of the observer with respect to the reference \ac{DRO} generated over a grid of $\alpha_h$.}\label{fig:DRO_single_target_relative_motion}
\end{figure}

In Figure~\ref{fig:DRO_single_target_absolute_motion} we plot the absolute motion of the observer trajectory in the rotating frame. On the left, we generate this for a homotopy $\alpha_h = 5\times10^{-3}$ and the right, $\alpha_h = 2\times10^{-2}$. In this example, the motion can be described totally in the $x-y$ plane. Blue and red portions of the trajectory correspond to coast and thrust stages, respectively. The black dashed line is the reference \ac{DRO}. While not immediately apparent, we notice that the right trajectory departs from the reference to a higher degree than the left. This effect is more apparent in Figure~\ref{fig:DRO_single_target_relative_motion}, where we plot the relative motion of the observer with respect to the reference \ac{DRO}. There, we clearly see that increasing $\alpha_h$ acts to provide a greater baseline distance over an observation interval. This behavior is expected because increasing the baseline generally provides a greater degree of dynamical differentiation, and thus, observability. This differentiation important for the \ac{DRO} family, which is fairly quiescent -- that is, this family does not exhibit the degree of nonlinearity associated with other periodic orbit families. 

\begin{figure}[hbt!]
\centering
\includegraphics[width=0.85\textwidth]{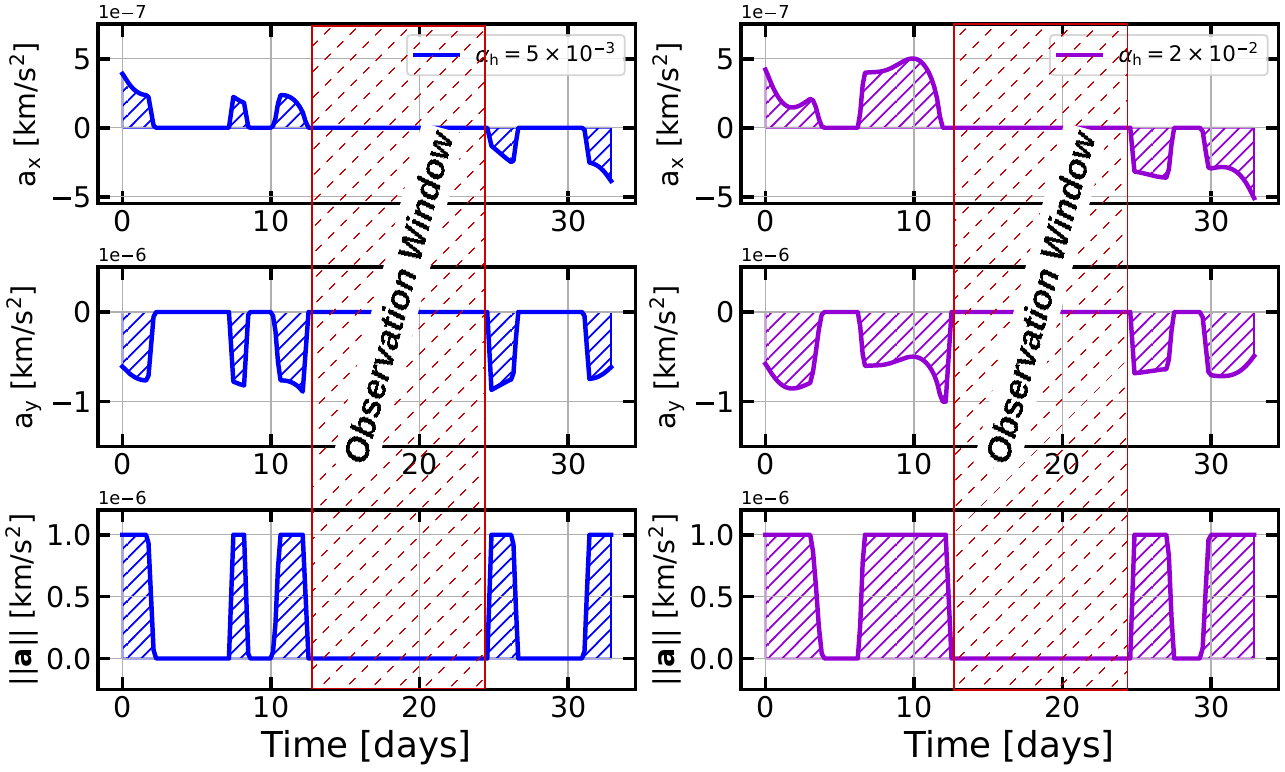}
\caption{The $a_x$, $a_y$, and thrust magnitude acceleration profile of the observer for $\alpha_h = 5\times10^{-3}$ (left), and $\alpha_h = 2\times10^{-2}$ (right).}\label{fig:DRO_single_target_thrust_curves}
\end{figure}
\begin{figure}[hbt!]
\centering
\includegraphics[width=0.85\textwidth]{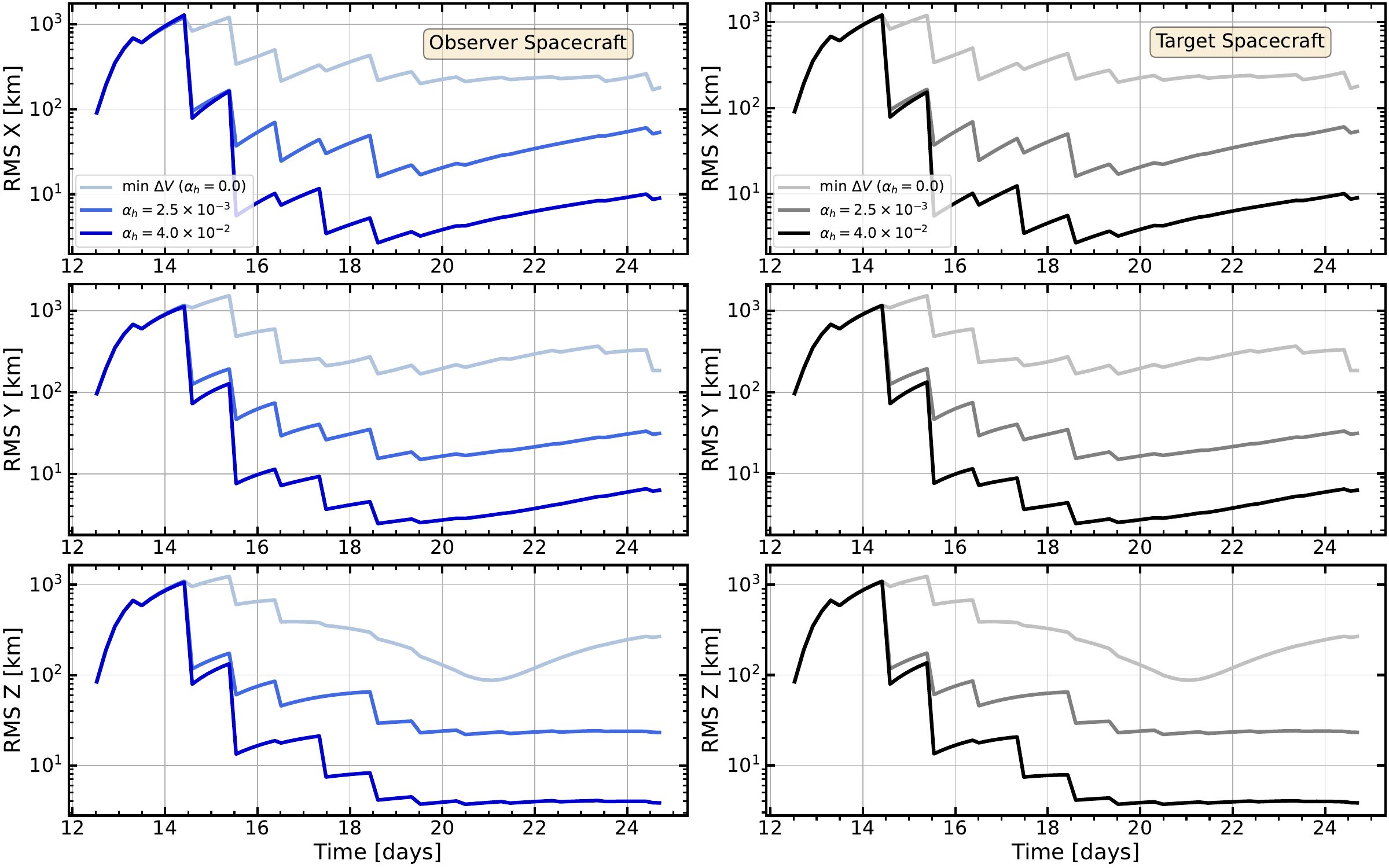}
\caption{\ac{CRLB} analysis for test case \#1. The left panels plot the predicted \ac{RMS} error in each position direction of the observer, and the right for the target.}\label{fig:DRO_single_target_CRLB}
\end{figure}
In Figure~\ref{fig:DRO_single_target_thrust_curves} we provide the thrust acceleration profile of the observer over the entire operating period. Again, the left panel corresponds to the profile for $\alpha_h = 5\times10^{-2}$ and the right for $\alpha_h = 2\times10^{-2}$. Increasing the homotopy accumulates a higher total impulse. For both values, in the observation window spanning between 12 and 23 days, zero thrust is commanded. The expected \ac{RMS} position errors, generated via a \ac{CRLB} analysis, are shown for in Figure \ref{fig:DRO_single_target_CRLB} for the observer (left) and target (right). There is a noticeable growth in predicted \ac{RMS} error at the start of the observation window, owing to our choice of error state covariance initialization -- recall, we use a diagonal covariance which does not provide state component correlations. Following the initial transient response, subsequent measurements serve to bound the expected error, but there is a substantial performance difference between homotopy values. By weighing the information gain to a higher degree, we can significantly reduce the expected RMS error for both the target and observer.

\begin{figure}[hbt!]
\centering
\includegraphics[width=0.75\textwidth]{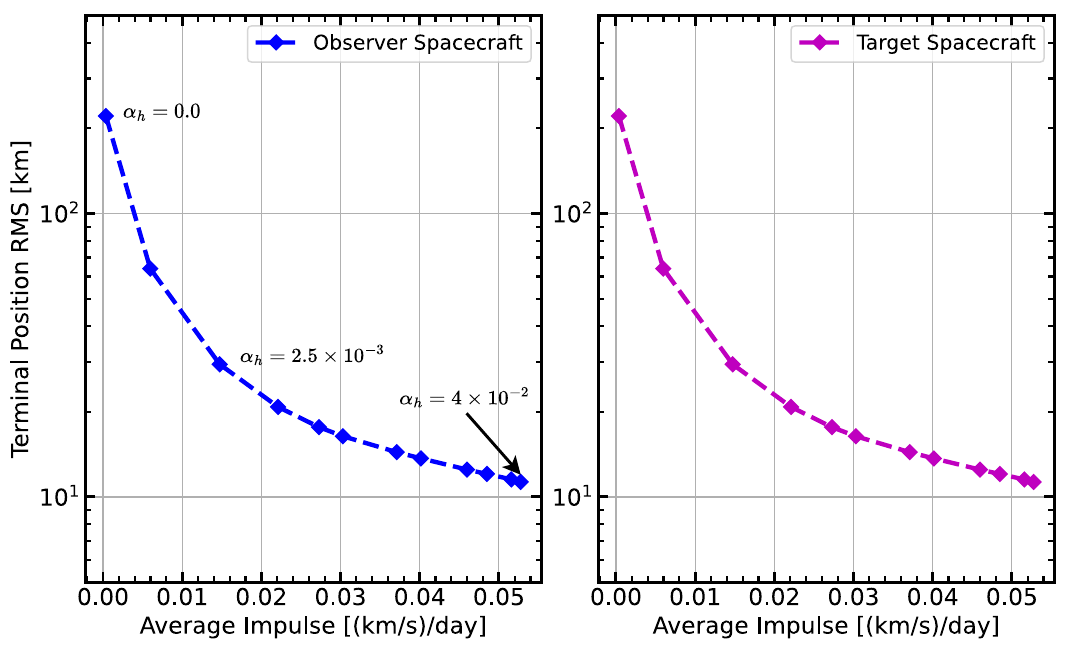}
\caption{Pareto curve of the average impulse and terminal expected RMS error for the observer (left), and target (right).}\label{fig:single_target_pareto}
\end{figure}

The trade-off between estimator performance and commanded control effort is most clearly shown in Figure~\ref{fig:single_target_pareto}. Here, we generate a set of optimal trajectories over a grid of $\alpha_h$ values, and plot both their corresponding average impulse against their predicted positional \ac{RMS} error at the conclusion of the observation window. Through this analysis, it is clear that the mutual information correlates well with the expected performance of a sequential estimator; albeit at the cost of greater control effort. This type of analysis also provides a design space of optimal trajectories, which may aid in mission design.  

\subsection{Test Case \#2}

Like the first test case, test case \#2 uses the same family of \acp{DRO}, but considers the joint estimation of the observer and three targets. Additionally, rather than relative positions, this case relies on range and range rate measurements, 
\begin{flalign*}
    \rho &= \sqrt{\Delta x^2 + \Delta y^2 + \Delta z^2},\\
    \dot{\rho} & = \Delta \dot{x} \frac{\Delta x}{\rho} + \Delta \dot{y} \frac{\Delta y}{\rho} + \Delta \dot{z} \frac{\Delta z}{\rho},  
\end{flalign*}
where, 
\begin{equation*}
    \Delta \boldsymbol{x} = \boldsymbol{x}_{\mathcal{T}} - \boldsymbol{x}_{\mathcal{S}}.
\end{equation*}
Measurements have an expected RMS error of 100 m in range and 10 m/s in range rate. The scenario duration is three periods of the reference \ac{DRO}, and, within this time span, the observation window is $[P_{\text{ref}}, 2 P_{\text{ref}}]$. Orbit characteristics for the observer and targets are contained in Table~\ref{tab:test_case_2_orbits}, and other relevant parameters in Table~\ref{tab:test_case_2_params}. 

\begin{table}[h!]
\caption{\label{tab:test_case_2_orbits} Orbit Parameters for Test Case \#2}
\centering
\setlength{\tabcolsep}{2pt}
\begin{tabular}{ccccccc}
\hline
\hspace{1ex} & $x_0$ [DU] & $y_0$ [DU] &$z_0$ [DU] & $\dot{x}_0$ [DU/TU] &$\dot{y}_0$ [DU/TU] & $\dot{z}_0$ [DU/TU]  \\\hline
$\boldsymbol{x}_{\mathcal{S}}(t_0)$ & $7.78185828\times10^{-1}$ & 0.0 & 0.0 & 0.0 & $5.28996986\times10^{-1}$ & 0.0 \\
$\boldsymbol{x}_{\mathcal{S}}(t_f)$ &  $7.80136159\times10^{-1}$ & 0.0 & 0.0 & 0.0  &$5.53104815\times10^{-1}$ & 0.0\\
$\boldsymbol{x}_{\mathcal{T}}^{(1)}(t_0)$ & $7.78717734\times10^{-1}$ & 0.0 & 0.0 & 0.0 & $5.55157488\times10^{-1}$ & 0.0 \\
$\boldsymbol{x}_{\mathcal{T}}^{(2)}(t_0)$ & $7.79426943\times10^{-1}$ & 0.0 & 0.0 & 0.0 & $5.54128887\times10^{-1}$ & 0.0 \\
$\boldsymbol{x}_{\mathcal{T}}^{(3)}(t_0)$ & $7.81554639\times10^{-1}$ & 0.0 & 0.0 & 0.0 & $5.51070308\times10^{-1}$ & 0.0 \\
\hline
\end{tabular}
\end{table}

\begin{table}[h!]
\caption{\label{tab:test_case_2_params} Scenario Parameters for Test Case \#1}
\centering
\begin{tabular}{lcccccc}
\hline
Parameter & & & Symbol & Value\\\hline
Initial Position RMS (Sensor) &&& $1\sigma$ [km] & 100 \\
Initial Velocity RMS (Sensor) &&& $1\sigma$ [km/s] & $1\times10^{-2}$\\
Initial Position RMS (Targets 1 -- 3) &&& $1\sigma$ [km] & 100\\
Initial Velocity RMS (Targets 1 -- 3) &&& $1\sigma$ [km/s] & $1\times10^{-2}$\\
Maximum Thrust Accel. &&& $a_{\text{max}}$ [km/$s^2$] & $1\times5^{-7}$ \\
Process Noise PSD  &&& $Q$ [km/$s^3$]  & $1\times10^{-11}$\\
Measurement Noise RMS (range) &&& $1\sigma$ [m]  & 100 \\
Measurement Noise RMS (range rate) &&& $1\sigma$ [m/s]  & 10 \\
Measurement Cadence &&& $f_{\text{meas}}$ [1/day] & 1.0 \\
\hline
\end{tabular}
\end{table}

\begin{figure}[hbt!]
\centering
\includegraphics[width=0.8\textwidth]{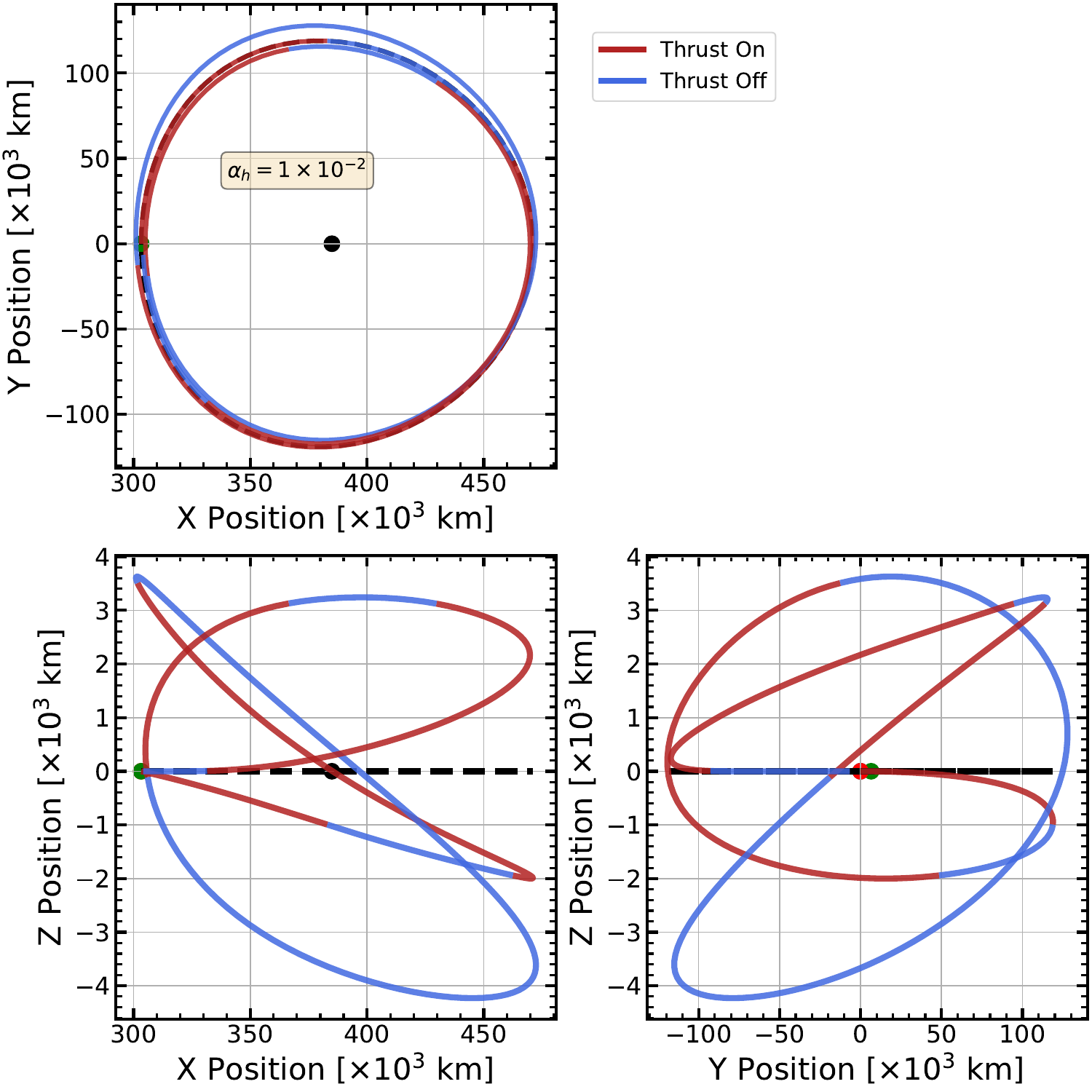}
\caption{Absolute trajectory of the observer for $\alpha_h = 1\times10^{-2}$. Each panel corresponds to the motion projected in one of the three principal planes.}\label{fig:DRO_multiple_target_absolute_motion}
\end{figure}

\begin{figure}[hbt!]
\centering
\includegraphics[width=0.8\textwidth]{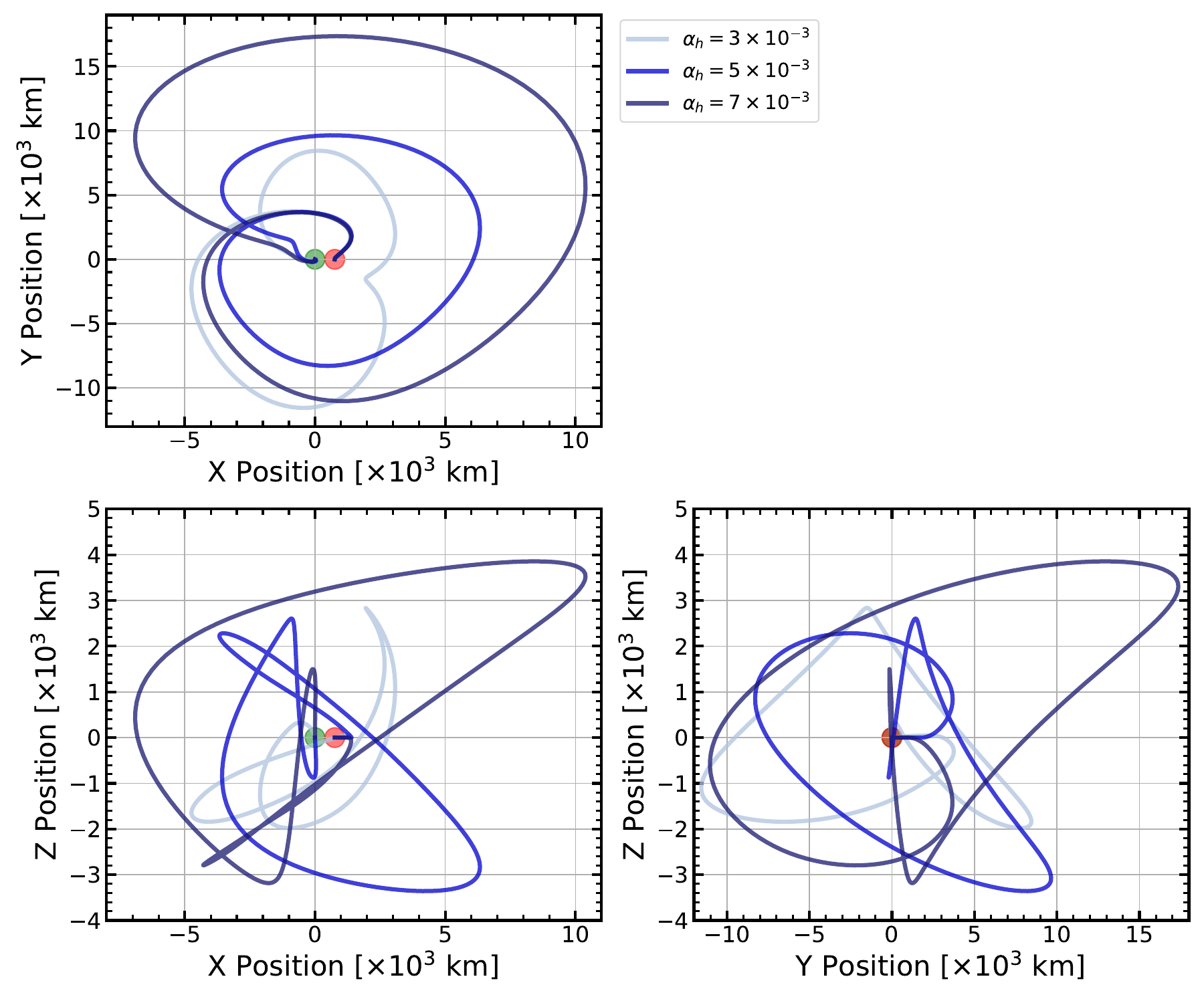}
\caption{Relative position of the sensor platform with respect to the reference \ac{DRO}. Here, we generate trajectories across a grid of $\alpha_h$. Each panel corresponds to the projected motion in one of the three principal planes. The green and red circles correspond to the initial and terminal positions.}\label{fig:DRO_multiple_target_relative_motion}
\end{figure}

The absolute motion of the observer over the scenario duration is shown in Figure~\ref{fig:DRO_multiple_target_absolute_motion} for a homotopy $\alpha_h = 1\times10^{-2}$. Unlike the previous case, the trajectory exhibits out of plane motion, so we show the trajectory projected onto each of the three principal planes. The out of plane motion can be attributed to the observability requirement for the range sensor; unlike the relative position measurements used in the previous case, the scenario requires motion in the $z$-axis direction. The blue and red segments once again correspond to coast and burn stages, respectively. The black dashed curves plots the reference \ac{DRO}. 

\begin{figure}[hbt!]
\centering
\includegraphics[width=0.8\textwidth]{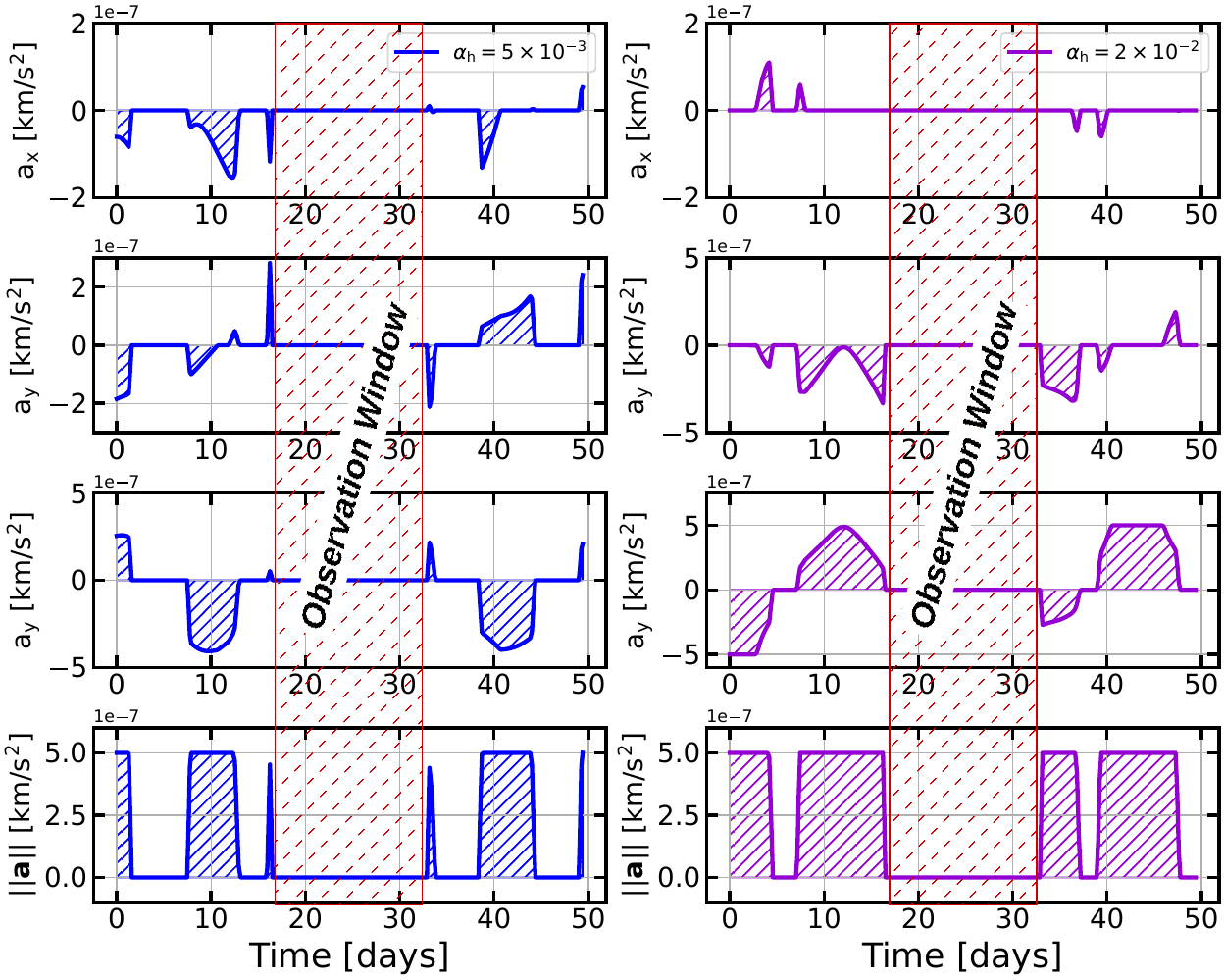}
\caption{The $a_x$, $a_y$, $a_z$, and thrust acceleration magnitude profiles of the observer for $\alpha_h = 5\times10^{-3}$ (left) and $\alpha_h = 2\times10^{-2}$ (right). }\label{fig:DRO_multiple_target_thrust_curves}
\end{figure}

The relative motion with respect to the reference orbit is plotted in Figure~\ref{fig:DRO_multiple_target_relative_motion}. Here, we generated trajectories across a grid of homotopy values and, as before, generally find that increasing $\alpha_h$ produces a path with a larger departure from the reference. This again is attributed to the gain in dynamical discrimination provided through a larger observer-target baseline distance. It is interesting to dissect some qualitative characteristics of the motion. For instance, in the $x - z$ and $y - z$ plane, while not exact, there is symmetry in the trajectory solutions for $\alpha_h = 5\times10^{-3}$ and $\alpha_h = 7\times10^{-3}$ across the $z$ axis. We do not provide a conclusion for this behavior, but postulate that these two trajectories may belong to a larger family of optimal solutions that exhibit symmetry like this. The green and red markers in the figure correspond to the initial and target states, respectively. 

\begin{figure}[hbt!]
\centering
\includegraphics[width=0.8\textwidth]{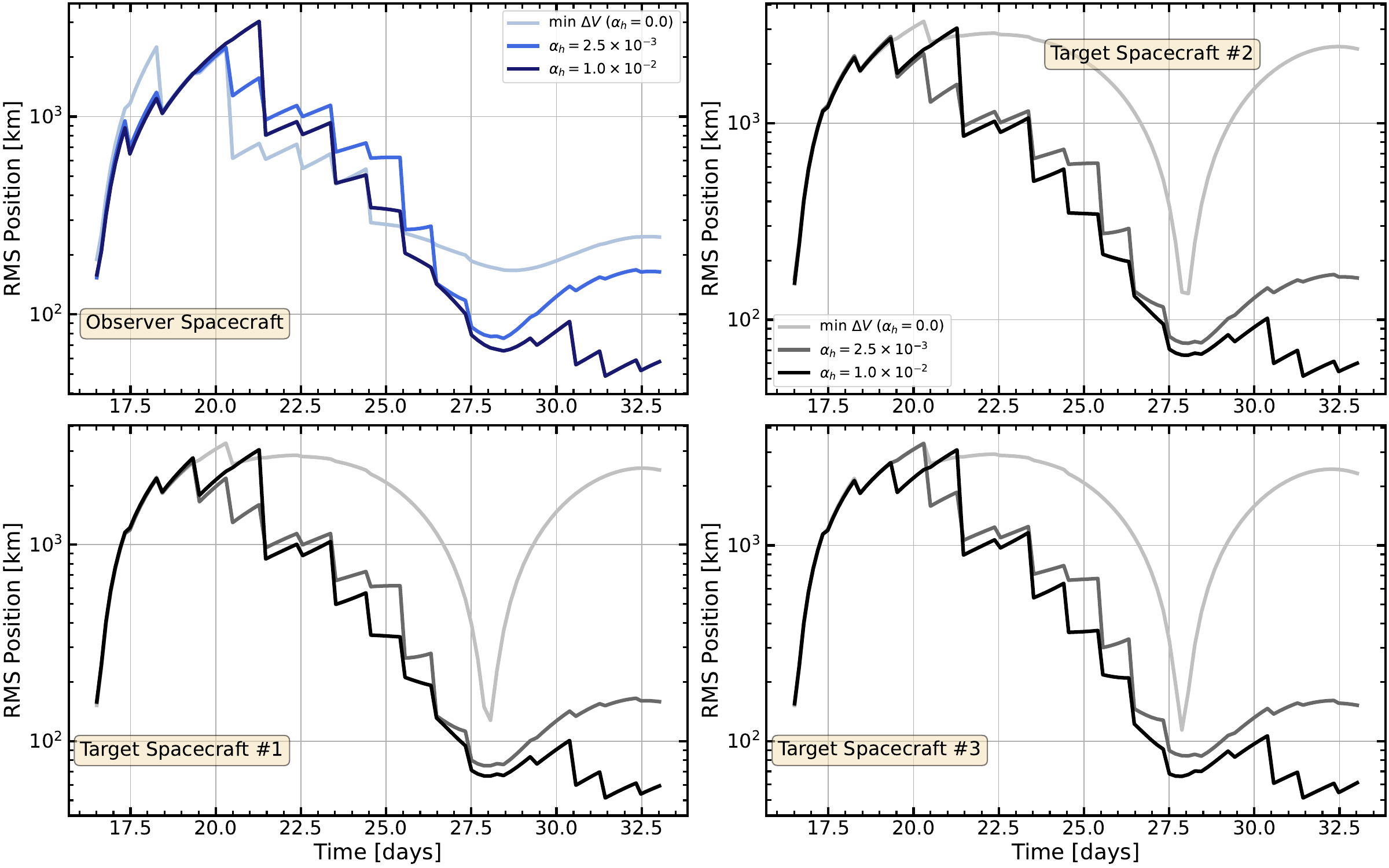}
\caption{Expected position \ac{RMS} error for the observer and each of the three targets.}\label{fig:DRO_multiple_target_CRLB}
\end{figure}

\begin{figure}[hbt!]
\centering
\includegraphics[width=0.75\textwidth]{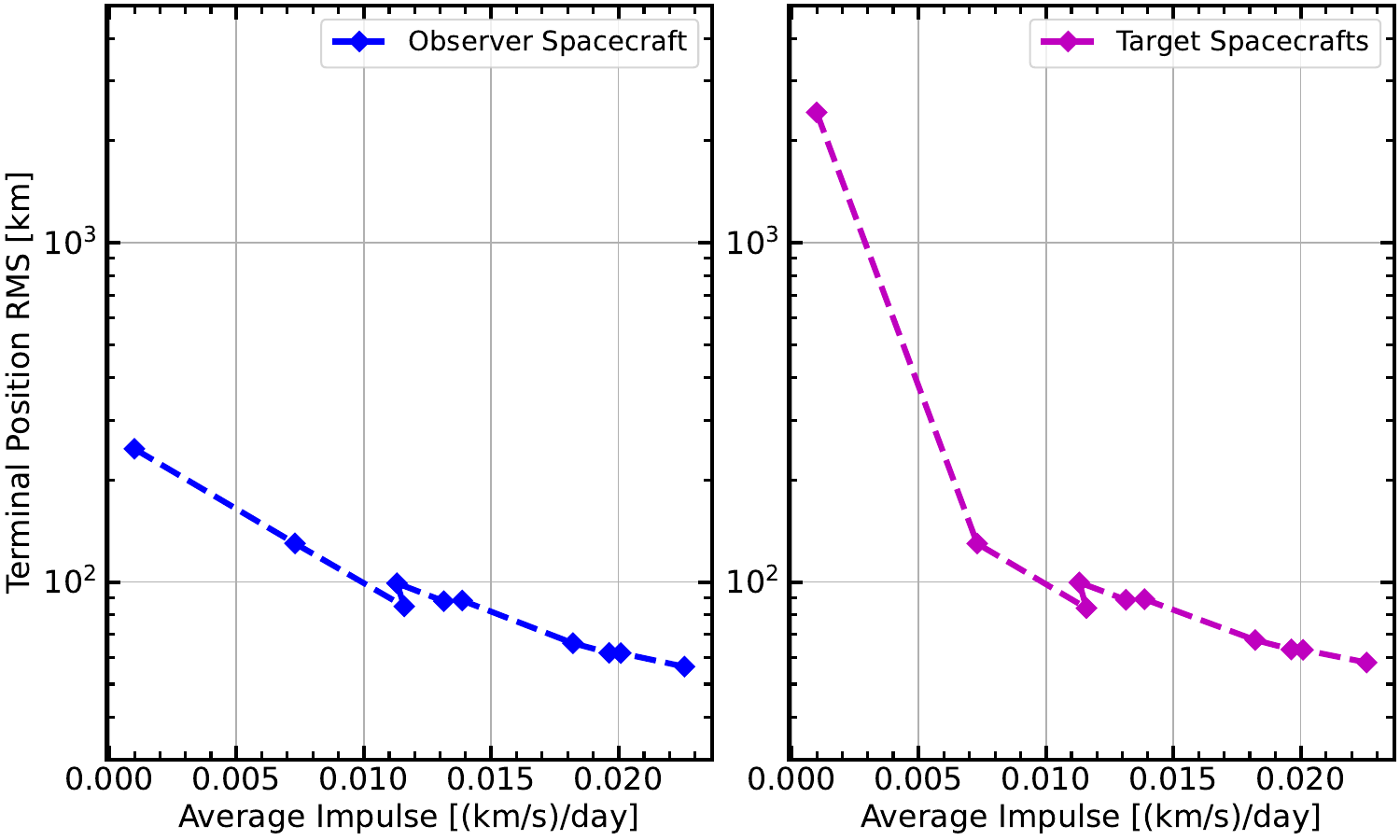}
\caption{Pareto curve of the average impulse and terminal expected position \ac{RMS} error for the observer (left) and targets (right).}\label{fig:multiple_target_pareto}
\end{figure}

Figure~\ref{fig:DRO_multiple_target_thrust_curves} plots the thrust acceleration profile in each direction, as well as the magnitude for $\alpha_h = 5\times10^{-3}$ and $\alpha_h = 2\times10^{-2}$. Unlike the previous test cases, this case commands control out of the $x-y$ plane. The window between roughly 17 and 32 days corresponds to the observation window, so the thrust acceleration in this portion is zero for both homotopy values. In Figure~\ref{fig:DRO_multiple_target_CRLB} we plot the expected RMS position error of a sequential estimator over the observation window, generated for both the observer and target. On average, the observer exhibits a smaller position \ac{RMS}, as it gains information from all three targets. In the minimum $\Delta V$ case (i.e., $\alpha_h = 0.0$) the system is weakly observable, and this is reflected in the light gray \ac{CRLB} analysis for each of the targets. Increasing $\alpha_h$ incentivizes information gain, and thus, produces trajectories that permit observability. The trade-off between estimator performance and control effort is shown in Figure~\ref{fig:multiple_target_pareto}. As before, it is clear that the mutual information correlates well with the expected performance of a sequential estimator.

%% file: conclusion.tex
In this study, we introduced a trajectory planning tool designed to maximize the navigation and tracking performance of a low-thrust observer operating in cislunar space, provided relative measurement of an object catalog. Our approach is based on maximizing a convex combination of the information gathered, generated through the mutual information, and the control effort exercised. In formulating this multi-objective optimization as an optimal control problem, we can generate a set of optimal solutions that lie along the Pareto curve of these two competing interests. The methods developed were applied to two relevant test cases of different measurement modality. In applying our methods, we show substantial improvement over a baseline comparison which utilizes solely control effort as an objective. Future work will consider alternate measurements, such as optical bearings, and more complex targeting and surveillance scenarios. 